\documentclass[11pt, a4paper, singlecolumn]{article}

\usepackage{color}

\usepackage[utf8]{inputenc}
\usepackage[english]{babel}

\usepackage[top=2cm, bottom=2cm, left=1.9cm, right=1.9cm]{geometry}

\usepackage[title]{appendix}

\usepackage{titlesec}
\titlespacing{\section}{0pt}{\parskip}{\parskip}
\titlespacing{\subsection}{0pt}{\parskip}{\parskip}
\titlespacing{\subsubsection}{0pt}{\parskip}{\parskip}

\usepackage{indentfirst}
\usepackage[export]{adjustbox}
\usepackage{graphicx}
\usepackage{amssymb}
\usepackage{amsmath}

\usepackage[dvipsnames]{xcolor}

\usepackage[figurename=Fig.]{caption}
\usepackage{caption}
\usepackage[colorlinks, citecolor=ForestGreen]{hyperref}
\usepackage{color, soul}
\usepackage{cite}

\usepackage{abstract}
\usepackage{multirow}
\usepackage[normalem]{ulem}
\usepackage{array}
\newcolumntype{*}{>{\global\let\currentrowstyle\relax}}
\newcolumntype{^}{>{\currentrowstyle}}

\title{\textbf{Semi-supervised Image Attribute Editing using Generative Adversarial Networks}\footnote{Note that this paper is the preprint of the accepted manuscript in Neurocomputing Journal. Download the high-quality version of the paper from this URL: \url{https://github.com/yahyadogan72/CRG}}}
\date{\vspace*{2pt}}
\author{
	\normalsize \textbf{Yahya Dogan}\\
	\normalsize Ankara University\\
	\normalsize Computer Engineering Department\\
	\normalsize yahyadogan@ankara.edu.tr
	\and
	\normalsize \textbf{Hacer Yalim Keles}\\
	\normalsize Ankara University\\
	\normalsize Computer Engineering Department\\
	\normalsize hkeles@ankara.edu.tr
}

\begin{document}
\maketitle

\begin{abstract}{
	\vspace*{-1.5em}
	\it Image attribute editing is a challenging problem that has been recently studied by many researchers using generative networks. The challenge is in the manipulation of selected attributes of images while preserving the other details. The method to achieve this goal is to find an accurate latent vector representation of an image and a direction corresponding to the attribute. Almost all the works in the literature use labeled datasets in a supervised setting for this purpose. In this study, we introduce an architecture called Cyclic Reverse Generator (CRG), which allows learning the inverse function of the generator accurately via an encoder in an unsupervised setting by utilizing cyclic cost minimization. Attribute editing is then performed using the CRG models for finding desired attribute representations in the latent space. In this work, we use two arbitrary reference images, with and without desired attributes, to compute an attribute direction for editing. We show that the proposed approach performs better in terms of image reconstruction compared to the existing end-to-end generative models both quantitatively and qualitatively. We demonstrate state-of-the-art results on both real images and generated images in CelebA dataset.
	
}\end{abstract}


\section{Introduction}

The aim of generative models is to produce synthetic images, which are similar to samples in a dataset. Recently, remarkable approaches have been proposed and developed in this field \cite{goodfellow2014generative, kingma2018glow, kingma2016improved, oord2016pixel}. Each of these approaches has their own strengths and weaknesses. Generative Adversarial Networks (GANs) \cite{goodfellow2014generative} dominate this field in terms of generating sharp images and having a meaningful latent representation with a rich linear structure \cite{radford2015unsupervised,rosca2017variational}. GANs are composed of two networks, i.e. generator and discriminator, that compete with each other during training for learning the underlying distributions of the images. Given a latent vector $z$, which is randomly drawn from normal or uniform distribution, namely $P(Z)$, the generator produces a synthetic image. The discriminator, on the other hand, determines whether a given image is real or fake. The discriminator produces a probability/score value that shows the probability/score of the given image as being a real image. Depending on the objective function that maximizes the discrimination of real and fake images, parameters of both networks, i.e. generator and discriminator, are optimized. It’s only after the Nash equilibrium that is established between the two networks, the discriminator hardly differentiates given images as real or fake with confidence. This state is a good indicator for the generator model that the generated images are similar to the real data distribution. After the training is completed, the discriminator is discarded and the generator is used for mapping from the latent space to the image space. 

Although a GAN generator provides a means for generation of different images using random latent vectors, it is difficult to model the inverse function for the generator model so that images can be manipulated in a controlled setting; especially for high resolution image generators. Existing methods apply algebraic operations in the latent vector space to encode semantically meaningful attributes to images.  In DCGAN work \cite{radford2015unsupervised}, algebraic operations have been performed on latent vectors of two generated images with and without an attribute, such as a face containing sunglasses or not; the latent vector representing the attribute is added to the latent vector corresponding to a generated image to provide the image with the attribute. However, due to use of randomly generated images, such vector additions manipulate other attributes as well as the desired attribute; hence usually cause significant changes in the original face attributes. The challenge here is to find a latent vector representation and a proper direction that corresponds to the factors that changes only the desired attribute; the produced images will be the same with the original source and they are different only for the encoded attribute. This operation requires accurate mapping from the image space to the latent space and identification of the true latent direction. In this work, we propose a novel architecture for solving both problems accurately. The proposed architecture is composed of an encoder and generator networks. The training of the networks are performed in an unsupervised setting, i.e. no attribute labels are used. Semi-supervision is performed in the calculation of desired attribute directions using the encoder network. We show that the proposed solution is effective for controlled manipulation of images.

The main contributions of this work can be summarized as follows: 

(1) We propose a new straightforward end-to-end architecture called Cyclic Reverse Generator (CRG) that allows learning the inverse of the generator in an unsupervised setting and allows reconstruction of both generated and real images in high quality. 

(2) We obtain a latent vector direction for attribute editing using our CRG encoder using only an arbitrary pair of real images; one with and the other without an attribute. 

(3) Keeping the other attributes the same, we show that it is possible to manipulate only the desired attributes at any rate. We provide an empirical analysis about determining the conservative manipulation ranges for different attributes.

(4) We give state-of-the-art results on CelebA dataset \cite{liu2015deep} to reconstruct an image at 128x128 resolution, which is the highest resolution that is obtained using an architecture with an encoder that is connected end-to-end to a generator.

The rest of the paper is organized as follows. A brief summary of the related previous works are summarized in Section \ref{sec:RelatedWorks}. The proposed method is discussed in detail in Section \ref{sec:TheMethod}. The results of our image reconstructions are provided in Section \ref{sec:experiments_and_results}. The paper is concluded with future directions.

\section{Related Works}
\label{sec:RelatedWorks}

\begin{figure}[!h]
	\centering
	\includegraphics[scale=.5]{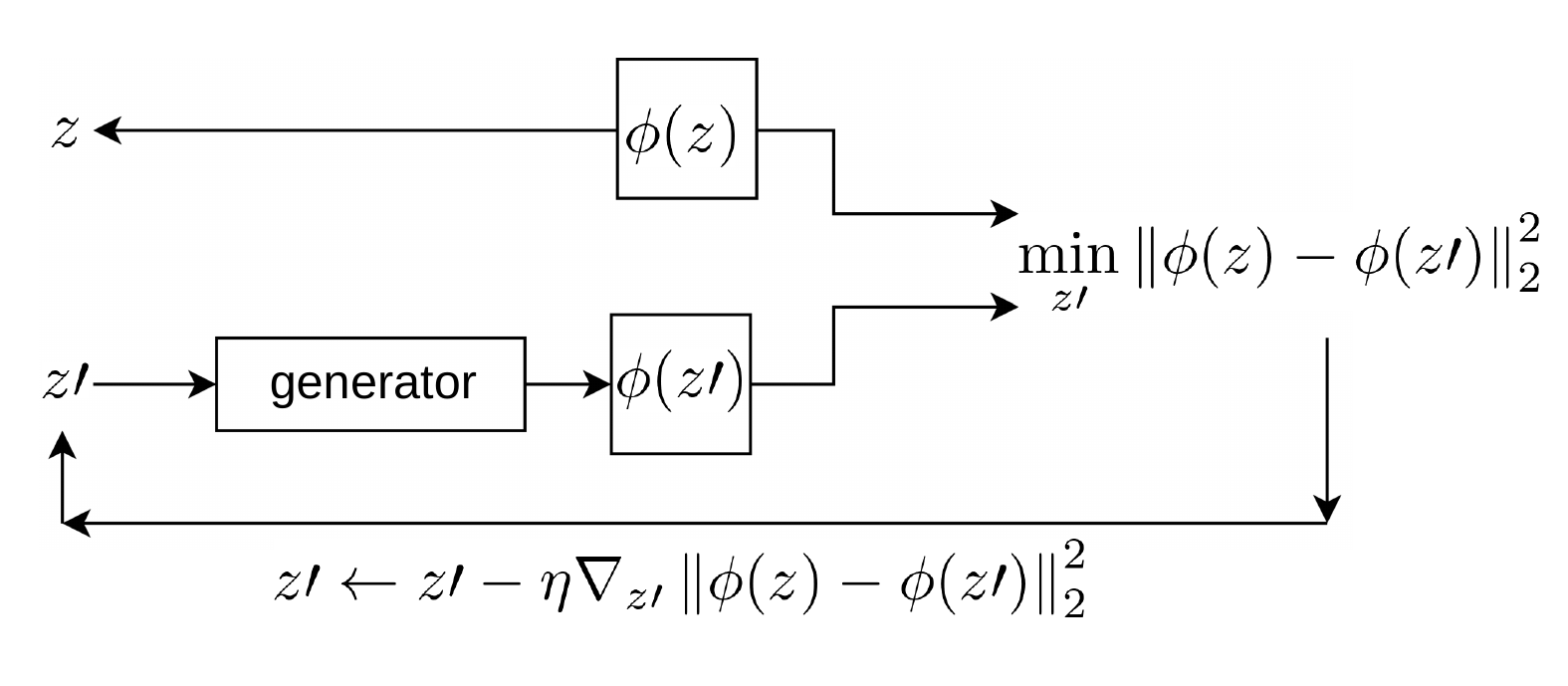}
	\centering
	\caption{The structure of the GBT \cite{lipton2017precise}.}
	\label{fig:GBT}
\end{figure}

 In GANs, the studies of manipulating desired attribute(s) of images can be divided into two main groups,  as semi-supervised and unsupervised methods. In the first group, training is performed using a code representing each attribute in a conditional GAN architecture \cite{mirza2014conditional, antipov2017face, perarnau2016invertible, jaiswal2018bidirectional, choi2018stargan,he2019attgan}. In these methods, after training, selected attributes of given images can be changed. However, the requirement of labeled datasets makes these methods impractical. Moreover, the attributes are limited with the trained labels.  In the second group, i.e. unsupervised methods. The main purpose is to construct a model that learns the hidden structures in image generation and to inverse the generation process without using attribute labels. Our work belongs to this group. The studies in this subgroup can be examined under three categories: The first one is Gradient Based Techniques (GBTs) \cite{lipton2017precise, creswell2018inverting}. In these approaches, latent vector corresponding to an image is taken as the optimization goal; the aim is to find latent vector $z\prime$  corresponding to $\phi(z)$  image. These methods do not require an additional encoder network. A pre-trained generator network is used to perform this process. Firstly, a $z\prime$ vector is sampled randomly from a prior distribution and it is given to the generator. According to the gradient obtained from the difference between the generated image and the target image, the $z\prime$ vector is updated. After numerous iterations, the $z\prime$ vector is expected to converge to the target vector $z$ that represents the $\phi(z)$ image. In GBTs, the encoding of the latent vector representation is obtained after a great number of gradient descent steps, which makes these approaches impractical. The structure of the GBT and the optimization function is depicted in Fig. \ref{fig:GBT}.

The second category includes architectures with an encoder network, in addition to the generator and the discriminator networks \cite{larsen2015autoencoding,donahue2016adversarial,dumoulin2016adversarially,luo2017learning}. These studies differ depending on how the encoder network is used and trained in the architecture. In the VAEGAN \cite{larsen2015autoencoding}, the generator is constructed using the decoder of a variational autoencoder (VAE), and encoder-decoder-discriminator networks are combined end-to-end as a single network. This structure enables reconstruction of more realistic and sharper images compared to VAEs. Similar to VAE-GAN, $\alpha$-GAN \cite{rosca2017variational} also combine variational and adversarial learning. This work also use additional discriminator network in the  latent space and additional pixel-wise reconstruction loss term in the image space. Unlike these two models, in BiGAN \cite{donahue2016adversarial} and ALI \cite{dumoulin2016adversarially} studies, neither generator nor encoder can see each other's outputs; in both works, the discriminator is trained to distinguish tuples of samples with their latent codes. In studies belonging to the second category, generally three networks are trained simultaneously. This makes it difficult to train encoder using a pre-trained generator. Another disadvantage of studies in this category is that they are inadequate to generate high-resolution synthetic images; most of these studies cannot go beyond 64x64 resolutions \cite{heljakka2018pioneer}. Moreover, most of these studies fail to extract the latent vector representation of an image that changes only a subset of attributes while preserving the other properties as they are. Another study in this category is the AEGAN architecture \cite{luo2017learning}. The difference of this architecture from the others is that the encoder network is not trained simultaneously with the other networks.  In this study, encoder-generator architecture is created by using a pre-trained generator model and an encoder network is trained from scratch. The encoder network is not trained with real images but is trained only with the generated images. The results of the reconstructions of the generated images are better than the real images; yet they are both not very successful.

The third category is Adversarial Generator-Encoder (AGE) architectures whose structure does not contain a discriminator network and an adversarial game is set up directly between the encoder and the generator \cite{ulyanov2018takes,heljakka2018pioneer}. In this category, training is performed using a combination of adversarial loss and reconstruction loss that encourages the encoder and generator to be reciprocal. There are three main advantages of these architectures: (1) Since adversarial loss is calculated between encoder and generator, there is no need for a discriminator and the number of learned parameters decreases, (2) In GANs, distributions are usually compared in high dimensional image space, in AGE, comparison is performed in a simpler latent space, (3) The encoder network is given both real and fake images as input, so the real images are better encoded. When outputs of this architecture are examined, it is seen that it produces better results than other end-to-end architectures for 64x64 resolutions, but it cannot go beyond 128x128 resolutions for reconstruction of images and the results at this resolution are not at desired level. Considering the studies in this category, it is seen that they are poor in terms of image quality and diversity compared to the models with a discriminator \cite{karras2017progressive}. Furthermore, if reconstruction loss is not used in AGE architecture, encoder and generator mapping is not reciprocal. In order to make networks reciprocal, reconstruction loss is controlled by a hyper-parameter in the objective function. However, the determination of the value of this parameter brings extra burden. To overcome the problems mentioned in this category, in this work we propose an architecture with two separate parts that have cyclic connections with each other and perform training in two steps. In the first step, a generator is trained using the original adversarial loss between the generator and discriminator \cite{karras2017progressive}. In the second step, the encoder network is trained from scratch by using the cyclic connection between the encoder and the generator. This architecture allows generator network to receive a latent vector, which is either randomly generated or encoded, as an input, and the encoder network to get either real or fake images as input during training. This approach helps convergence problems of encoders during training. Moreover, it has an additional benefit that it enables learning the inverse mapping of any pre-trained generator that are readily available for generation of images in different domains.

The success of the encoder network is directly related to the success of the generator in mapping latent vectors to images. Two main problems arise in generator training, especially when high-resolution images are used. The first problem is the stability problem; competition between the two networks, i.e. generator and discriminator, creates instability, since one dominates the other. The second problem is the mode collapse problem; the generator network produces samples in limited varieties. Recently, new objective functions \cite{arjovsky2017wasserstein,gulrajani2017improved,kodali2017convergence,mao2017least,berthelot2017began}, regularization techniques \cite{gulrajani2017improved,miyato2018spectral} and network architectures \cite{radford2015unsupervised,berthelot2017began,karras2017progressive} have been proposed to overcome the stability problem. In order to prevent mode collapse problem, generating similar images are prevented using similarities of the samples in the discriminator networks \cite{salimans2016improved,karras2017progressive}. Although there is various GAN research that generate high quality images for popular datasets, it is not clear which algorithm is superior to the others. In a recent study, the state-of-the-art models have been compared objectively using some well-known evaluation metrics \cite{lucic2018gans}. In this study, it is reported that the performance of the models largely depends on datasets and hyper-parameter optimization; there is no ideal model that consistently generate high quality images in all datasets. Therefore, it is a challenge to determine the appropriate architecture, objective function and regularization techniques for different datasets.

In addition to the vast literature about GAN training, there are recent works that focus on generating images with different attributes by encoding those attributes as part of the latent code \cite{mirza2014conditional, antipov2017face, perarnau2016invertible, jaiswal2018bidirectional, choi2018stargan,he2019attgan}. Our work is in a different line from these works; we do not explicitly represent attribute codes during training. These works simply utilize the attribute codes in the latent vectors to edit a particular attribute of an image. Yet, the attribute editing is still a challenging problem without using explicit attribute codes, since the encoding of the desired attributes in the latent space is entangled. Therefore, it is difficult to change only a subset of attributes without changing other properties of the image. Recently, different from the existing approaches, StyleGAN \cite{karras2018style} work presents an unsupervised approach to learn relevant attribute directions. In the StyleGAN, research has been carried out on making the latent vector disentangled by using the generator structure used in the style transfer literature \cite{huang2017arbitrary}. The aim is to find a latent vector that is composed of linear subspaces, such that each subspace controls an attribute. With this architecture, low, medium and high level attributes in the image can be modified by changing the determined subspaces of the latent space corresponding to the image. Still yet, the changes in this subspace changes other attributes of the image as well as the desired attributes.

\section{The Method}
\label{sec:TheMethod}

In the proposed method, we first trained a generator model in a GAN setting and used that pre-trained generator network to train an encoder that learns the inverse mapping of the generator. We propose a new cyclic parameter optimization for learning this inverse mapping. The mapping from the image space to the latent vector space is used actively to identify relevant attribute directions for image attribute editing. We show the efficacy of our proposal using the CelebA dataset. 

This section is organized as follows. We first introduce the proposed Cyclic Reverse Generator (CRG) Model, which enables reconstruction of images via a cyclic error minimization function. Following that, we provide the training details of our generator and encoder networks using CelebA dataset. Then, we explain how we compute the latent attribute representations with a pair of reference images using the CRG model. Finally, we provide empirical analysis of the computed attribute directions to show that the computed attribute directions are relevant to the target attributes.

\subsection{The Cyclic Reverse Generator (CRG) Model}

\label{sec:CRG_model}

\begin{figure}[!h]
	\centering
	\includegraphics[scale=.8]{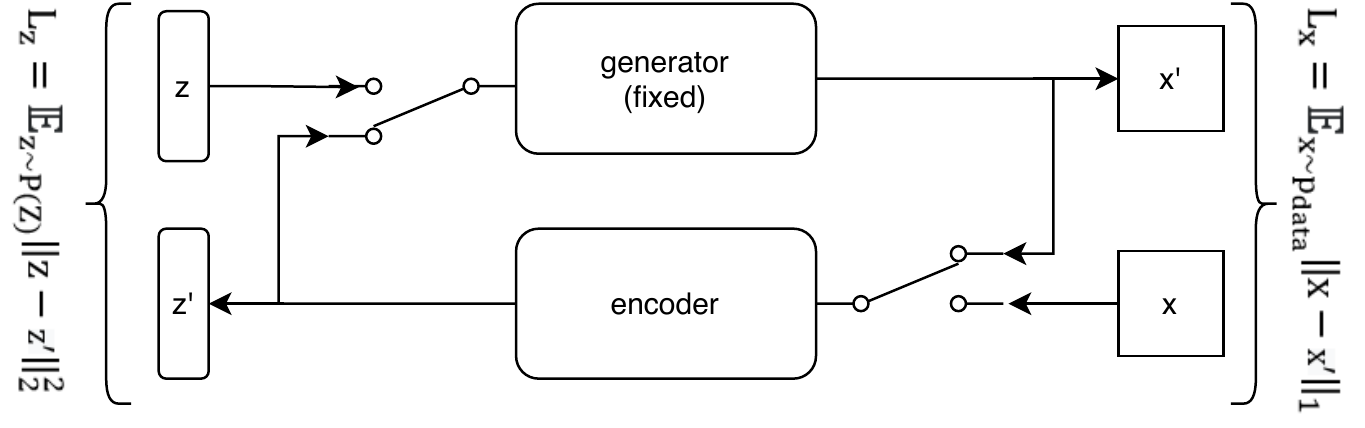}
	\centering
	\caption{The CRG model.}
	\label{fig:CRG}
\end{figure}

This section introduces the proposed model for the reconstruction of an image. Our model is composed of two parametric networks: the generator network $g_{\omega}(z)$ that maps the latent space $z$ to the data space $x$, and the encoder network $e_{\epsilon}(z)$ that maps the image $x$ from the data space to the latent space $z$. The goal is to ensure that the encoder and the generator networks have bidirectional connections for generating and encoding of either real or fake images. The design of the proposed model is shown in Fig. \ref{fig:CRG}. The reconstruction loss is calculated using the error in the image space (using $L_{x}$, in Fig. \ref{fig:CRG}) and the error in the latent space (using $L_{z}$, in Fig. \ref{fig:CRG}) together, in an order, during encoder training.

We first train the GAN architecture, i.e. generator and discriminator networks, until we obtain state-of-the-art scores for the selected datasets. We observed that, it is only after an optimum generator is trained that a successful mapping can be done from the image space, $x$, to the latent space, $z$. Our first expectation from the model is the ability to reconstruct the latent representation of randomly generated samples.  For a given random $z$, the generator generates an image, $x^\prime$; giving that image to the encoder network an estimated $z$ is obtained, i.e. $z^\prime$. We minimize the objective in (\ref{egu:z_optimize}) to minimize the error between the estimated $z^\prime$, i.e. $e_{\epsilon}(g_{\omega}(z))$, and the initial $z$.

\begin{equation}
\label{egu:z_optimize}
\hat{\epsilon} =\operatorname*{argmin}_{\epsilon}(\mathbb{E}_{z \sim P(Z)}\left\|z-e_{\epsilon}(g_{\omega}(z)) \right\| _{2}^2)
\end{equation}

Our second expectation from the model is the ability to reconstruct real images. This is also important because we want to be able to compute the relevant directions related to an attribute in the latent $z$ space using real reference images. For a given image $x$, the encoded $z$ value, $z^\prime$, is used to reconstruct the image $x^\prime$, i.e. $g_{\omega}(e_{\epsilon}(x))$, we use the objective in (\ref{egu:x_optimize}) to minimize the error between the reconstructed image $x^\prime$ and the original image $x$.

\begin{equation}
\label{egu:x_optimize}
\hat{\epsilon} =\operatorname*{argmin}_{\epsilon}(\mathbb{E}_{x \sim P_{data}}\left\|x-g_{\omega}(e_{\epsilon}(x)) \right\| _{1})
\end{equation}

In each iteration, the parameters of the encoder are updated twice, using the objective in (\ref{egu:z_optimize}) and (\ref{egu:x_optimize}), respectively. 

We call our proposed architecture as Cyclic Reverse Generator (CRG) (Fig. \ref{fig:CRG}), since the encoder learns the inverse function of the generator using the proposed cyclic error minimization. In the model, the encoder is given either real or generated images, and the generator is given either a randomly generated or an encoded latent vector as the input. This enables both generated and real images to be reconstructed. We use a pre-trained generator in the CRG architecture. In this setting, only the encoder parameters are updated, the generator is fixed. We also included a discussion with a non-fixed generator in Section \ref{sec:image_reconstruction_experiments}. Generally, end-to-end approaches implement simultaneous training of three networks, i.e. generator, discriminator and encoder networks, since the dominance of one of the networks in the architecture usually leads to instability problems. Therefore, a pre-trained generator is not used in these studies. However, in CRG, the encoder is trained utilizing a pre-trained generator without facing stability problems. This is advantageous since we can train an encoder for a GAN model that is already trained for different domains.

\subsection{Training The Models}
\label{sec:Training_the_models}
In this section, we provide the training details for the GAN and the encoder models using the CelebA dataset.

Recently, a lot of researchers are working actively in the GAN domain and it is a challenge to determine which architecture(s) and objective function(s) are better suited for training a GAN for a particular dataset. Authors of \cite{lucic2018gans} argue that the performance of recently proposed state of the art models heavily depends on datasets, and that no particular model is strictly dominating the others. In this work, we used a slightly modified version of the Progressive GAN (PGGAN) architecture for GAN training, since it generates high quality images\cite{karras2017progressive}.

\subsubsection{GAN Model for CelebA Dataset}
\label{sec:CelebAModel}
The CelebA dataset contains $202599$ celebrity images with large pose variations, taken from different backgrounds. In this study, we use $30000$ images in 128x128 pixel resolution. Most of the models that infer the latent $z$ vectors work with 32x32 resolution images. There are very few models that go beyond 64x64 pixel resolution \cite{heljakka2018pioneer}. Current state-of-the art models in the $z$ inference produce a maximum of 128x128 images. However, the quality of the images is not sufficiently good for the reconstruction of the real images that are necessary to find the latent vector corresponding to an attribute. The reason for this is that as image resolution increases, stability and diversity problems become more apparent in GAN training. Hence, in order to achieve the purpose of this research, we need to train a generator that enable us to generate high-resolution images without encountering stability and diversity problems.

\begin{figure}[!h]
	\centering
	\includegraphics[width=0.97\textwidth]{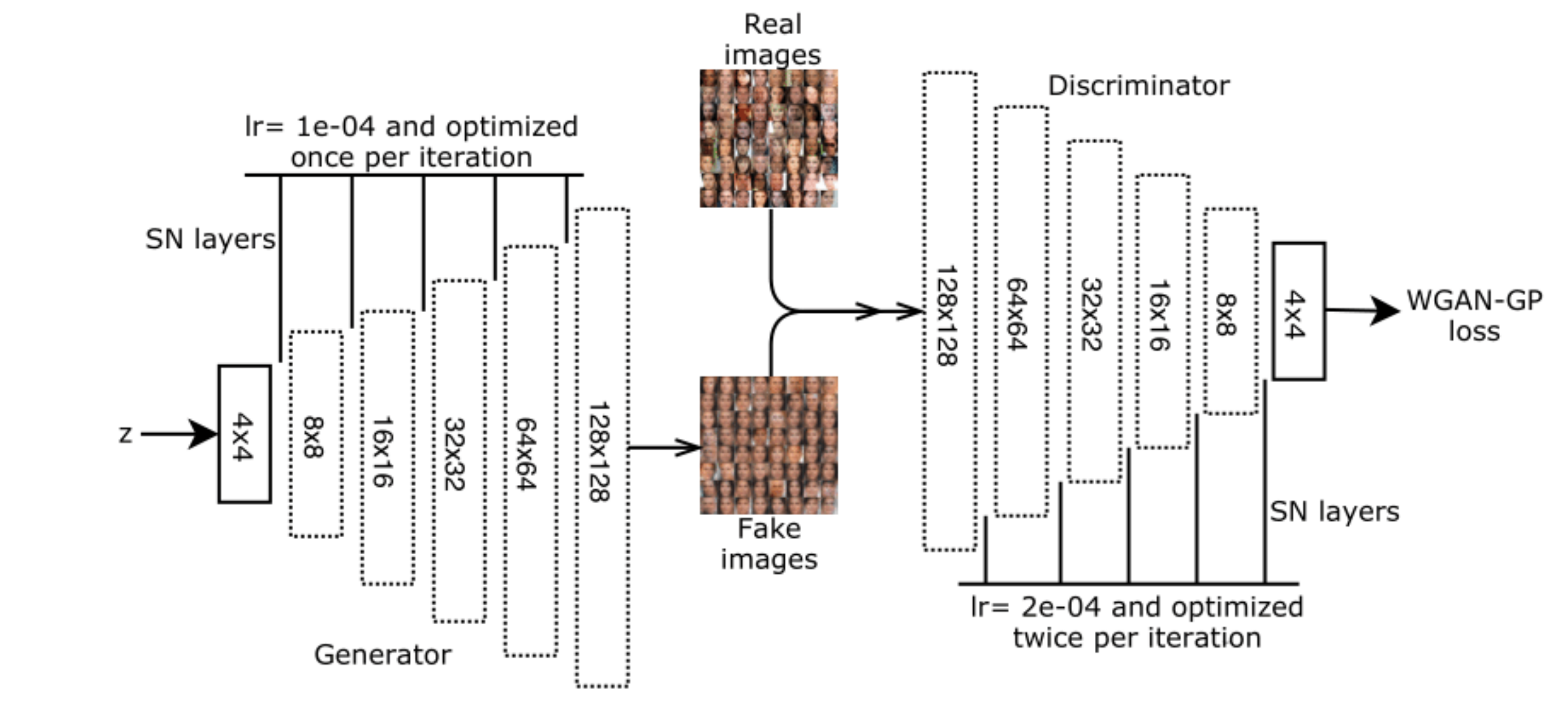}
	\centering
	\caption{The GAN architecture for CelebA model.}
	\label{fig:GAN_arch}
\end{figure}

\paragraph{The Generator Network:}
There are some design choices to be made when training a generator network to generate high-resolution images; selection of the objective function, used regularization technique, selected GAN architecture and selection of the hyper-parameter values. For the CelebA dataset, we use the same model that we proposed in our preliminary work \cite{doganyahya}. Our model is similar to the PGGAN model, with some additional normalization layers; the architectural details are provided in Fig. \ref{fig:GAN_arch}.  

In GAN studies, as the image resolution increases, the discriminator distinguishes between real and generated images easily. This causes imbalance problem between the two networks during training. The key idea in PGGAN is to grow both the generator and the discriminator progressively to capture fine details in images. Initially, the training of both networks is started with a low spatial resolution, i.e. 4x4 pixel images, and as the training progresses, a new layer is added to both networks to increase the spatial resolution of the generated images. Differently from the PGGAN, to ensure the stability between the two networks, we apply spectral normalization \cite{miyato2018spectral} to both the generator and the discriminator as in SAGAN \cite{zhang2018self}, we keep the learning rates of the discriminator more than the generator as in TTUR \cite{heusel2017gans}, i.e. we use $1e-4$ and $2e-4$ learning rate for the generator and the discriminator, respectively. We optimize the generator and the discriminator by 1:2 rate. 

\begin{table}[!h]
	\caption{Performance comparison of the generator networks for 128x128 pixel resolution.}
	\centering
	\label{tab:table2}
	\begin{tabular}{ccc}
		\hline
		& FID \cite{heusel2017gans}            & \begin{tabular}[c]{@{}c@{}}Latent vector\\ dimension\end{tabular} \\ \hline
		AGE \cite{ulyanov2018takes}                                 & 154.79        & 512                                                               \\ \hline
		Pioneer \cite{heljakka2018pioneer}                              & 23.15         & 512                                                               \\ \hline
		\multirow{2}{*}{\textbf{Our model}} & \textbf{8.96} & \textbf{512}                                                      \\ \cline{2-3} 
		& \textbf{9.4}  & \textbf{128}                                                      \\ \hline
	\end{tabular}
\end{table}

\begin{figure}[!h]
	\centering
	\includegraphics[width=.9\textwidth]{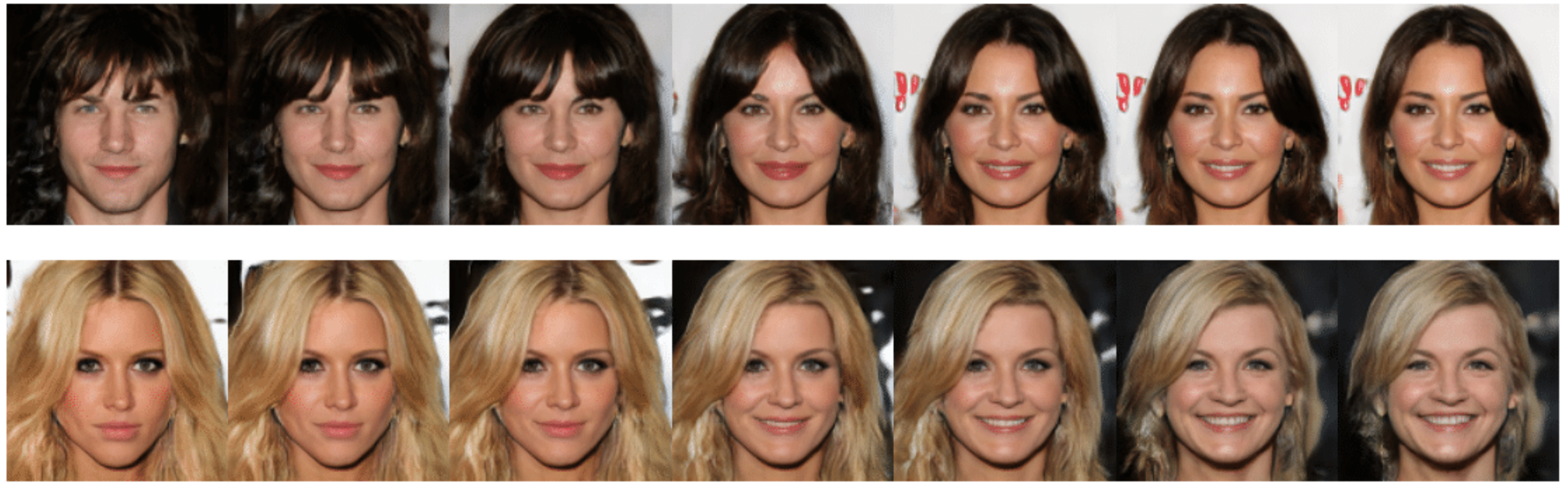}
	\caption{Sample interpolations of our generator between two generated images.}
	\label{fig:interpolation}
\end{figure}

In Table \ref{tab:table2}, we compare our generator with the AGE and Pioneer \cite{heljakka2018pioneer} generators for 128x128 pixel resolution, since these models contain a model that infer the latent $z$ vectors similar to ours. Pioneer is the progressive model of the AGE architecture. We use the Fréchet Inception Distance (FID) metric, which is a commonly used metric to compare generator networks in terms of image quality and diversity. A low FID score indicates a better generator. Note that our generator produces lower FID scores compared to the other methods, for both $128$ and $512$ latent vector dimensions. In the training of our CRG model, we preferred the generator network with lower dimension, i.e. $128$, since there is no significant margin in the FID scores between $128$ and $512$. The results also depict that using a model with a discriminator produces better scores. In Fig. \ref{fig:interpolation}, we show the interpolation capability between randomly generated images using our generator. The quality of the generated images is apparent in these randomly generated samples.  

\subsubsection{The Encoder Model}
\begin{figure*}
	\centering
	\includegraphics[width=.9\textwidth]{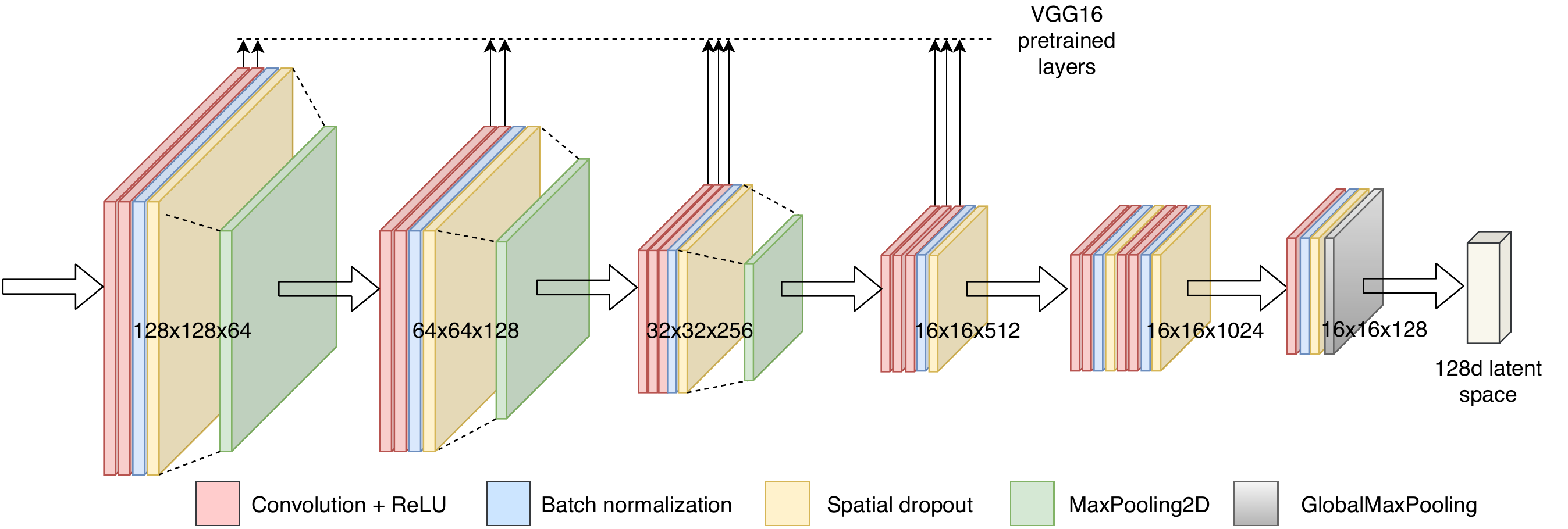}
	\caption{The Encoder network architecture.}
	\label{fig:encoder}
\end{figure*}

We designed our encoder model as a multi-layer convolutional neural network. The generator model described in the previous section contains approximately $22$ million trainable parameters. In general, we observe that keeping the capacity of the encoder higher than the generator provides better encoding. The encoder consists of $6$ blocks with a total of approximately $26$ million trainable parameters. Training deep CNNs from scratch can cause convergence problems during the network training and training time increases substantially \cite{zeiler2014visualizing}. Given the fact that CNNs encode low-level features in the first few layers, we use the first $4$ blocks of the pre-trained VGG-16 Net \cite{simonyan2014very} in the encoder architecture with a list of modifications: (1) we remove the fourth max-pooling layer, so as not to reduce image resolution too much, (2) we include a batch normalization layer at the end of each convolution block to overcome the vanishing gradient problem that occurs when we include additional blocks at the end, (3) to avoid overfitting, we use spatial dropout layer after the batch normalization layers. In order to increase the capacity of the network, we include two new blocks to extract features that are specific to CelebA dataset. Finally, we use global max pooling on the last layer of the encoder. There are $64$ filters in the first block and the number of filters in the consecutive blocks are doubled except for the last block; since the latent vector dimension is $128$, we use $128$ filters in the last block. We use a 3x3 filter size in the convolution layers, except for the 1x1 transposed convolution layer in the last convolution layer. The structure of the encoder is depicted in Fig. \ref{fig:encoder}.

\paragraph{Training details:}
We trained the encoder using the CRG architecture as we explained in Section \ref{sec:CRG_model}. As previously mentioned in the CRG, MAE loss is used in the image space and MSE loss is used in the latent space, and only the parameters of the encoder are optimized at each iteration. During the training, all of the weights of the convolutional layers of the encoder, including the pretrained weights of the VGG Network, are trained using the same learning rate. We train the encoder using RMSProp optimizer \cite{tieleman2012lecture} with a batch size of $128$, setting rho to $0.9$ and epsilon to $1e-08$. In order to adapt the existing VGG layer weight values properly, we use a small learning rate, i.e. $1e-4$, for optimizing all the parameters in the encoder architecture. We reduce the learning rate by a factor of $2$ when minimum validation loss stops improving for $10$ epochs. To avoid overfitting, we apply different regularizations in the training; we use 50\% spatial dropout, and data augmentation where we rotate the images in the range of $30$ degrees and apply horizontal and vertical flips. Model checkpoint is used to save the best model and using early stopping, we end the training process if no improvement is achieved in the validation loss for $20$ epochs. After the training process, we expect that the generated images and their reconstructions look almost the same, and that the reconstructions of the real images are at an acceptable level; sufficient to find accurate latent vector directions for the attribute embedding. 

\subsection{Computing Latent Attribute Representations}
\label{sec:computing_latent_attribute}
\begin{figure}[!h]
	\centering
	\includegraphics[width=.5\textwidth]{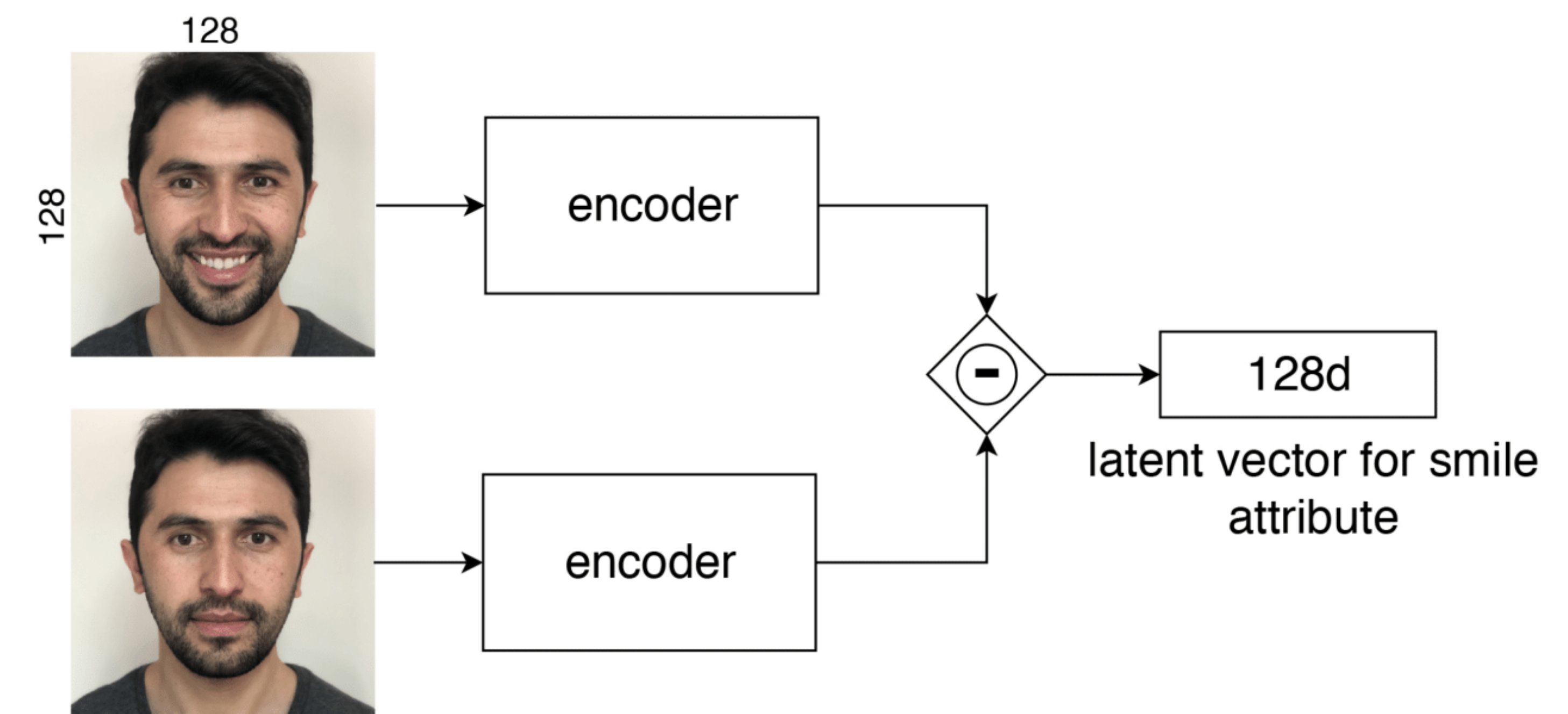}
	\caption{Extraction of the latent vector corresponding to smile attribute.}
	\label{fig:smile_attribute}
\end{figure}

In the CRG architecture, the encoder is trained completely without any supervision; we do not use any labels or encode conditions during GAN training (please refer to Section \ref{sec:Training_the_models} for details). In order to extract the latent vector corresponding to an attribute, we use an arbitrary real image as a reference and a second image that contains a particular attribute that we want to control. In Fig. \ref{fig:smile_attribute}, we demonstrate the extraction of the latent vector direction for smile attribute embedding. We first use our encoder to get the latent vectors of the reference images. Then compute the normalized direction in the latent space to determine the relevant attribute direction.

When the accuracy of the generator and its inverse function, i.e. the encoder, is high, this direction accurately changes only the relevant attribute by preserving the other attributes in the image. Equation (\ref{equ:feature_normalized}) is used to find the latent vector direction representing an attribute $f$.

\begin{equation}
\label{equ:feature_normalized}
z_{f}=\frac{z_{2}-z_{1}}{\left\|z_{2}-z_{1} \right\|_{2}^{2}}
\end{equation}

Where $z_{1}$ is the latent code that is generated by the encoder using a reference image and $z_{2}$ is the latent code obtained using an image that has the desired attribute. Note that, depending on the context, attributes can be encoded in both positive or negative directions, i.e. from  blond hair to brown hair ($z_{1}-z_{2}$) or brown hair to blond hair ($z_{2}-z_{1}$); then $z_{1}$ and $z_{2}$ in Eq. (\ref{equ:feature_normalized}) needs to be configured accordingly. After $z_{f}$ is obtained, equation (\ref{equ:z_new}) is used to generate an image with the desired attribute.

\begin{equation}
\label{equ:z_new}
z_{a}=z_{p}+k*z_{f}
\end{equation}

Where $z_{p}$ is the latent vector estimated by the encoder using any arbitrary image, k is the amount of attribute that we want to impose into the image, $z_{a}$ is the latent code of the attribute encoded image. The attribute encoding can be performed in positive or negative directions, as desired, by changing the sign of k accordingly in (4). To the best of our knowledge, the existing reconstruction models using GANs are not yet capable of accurately controlling only selected attributes in images. Modifying an image this way has three main advantages: (1) there is no need to have a labeled dataset to encode attributes, (2) attribute set is not constant; it can be constructed anyhow as long as a pair of arbitrary reference images are provided, (3) the amount of attribute injection to the target image can be controlled easily by adjusting one parameter, i.e. $k$.


\subsection{Analysis of Attribute Directions ($z_{f}$):}
\label{sec:Analysis_of_attrib_dir}

In this section, we provide an empirical evaluation of a set of attributes using randomly selected real images that are unseen by our GAN training with and without the selected attributes from the CelebA dataset. Note that, these images have attribute labels in the CelebA dataset; we use these labels to easily analyze the relative positions of the images with and without attributes on the projection axis. For each attribute\footnote{For the eyeglasses attribute, due to the lack of enough samples, we use around $13000$ images.}, we selected around $20000$ neutral images, i.e. images without the selected attribute, and $20000$ images with the attribute. We want to observe the relative positions of the neutral and attributed images in the latent space by projecting them to the computed attribute vector directions for four attributes: smile, hair color, eyeglasses and beard. According to our proposal, an image without an attribute changes gradually to contain the attribute when we iterate the latent vector of the image in the computed latent attribute direction; hence we expect to see two different modalities on the projection line when we project the neutral and the attributed sample images on this line. 

\begin{figure}[!h]
	\centering
	\includegraphics[width=.9\textwidth]{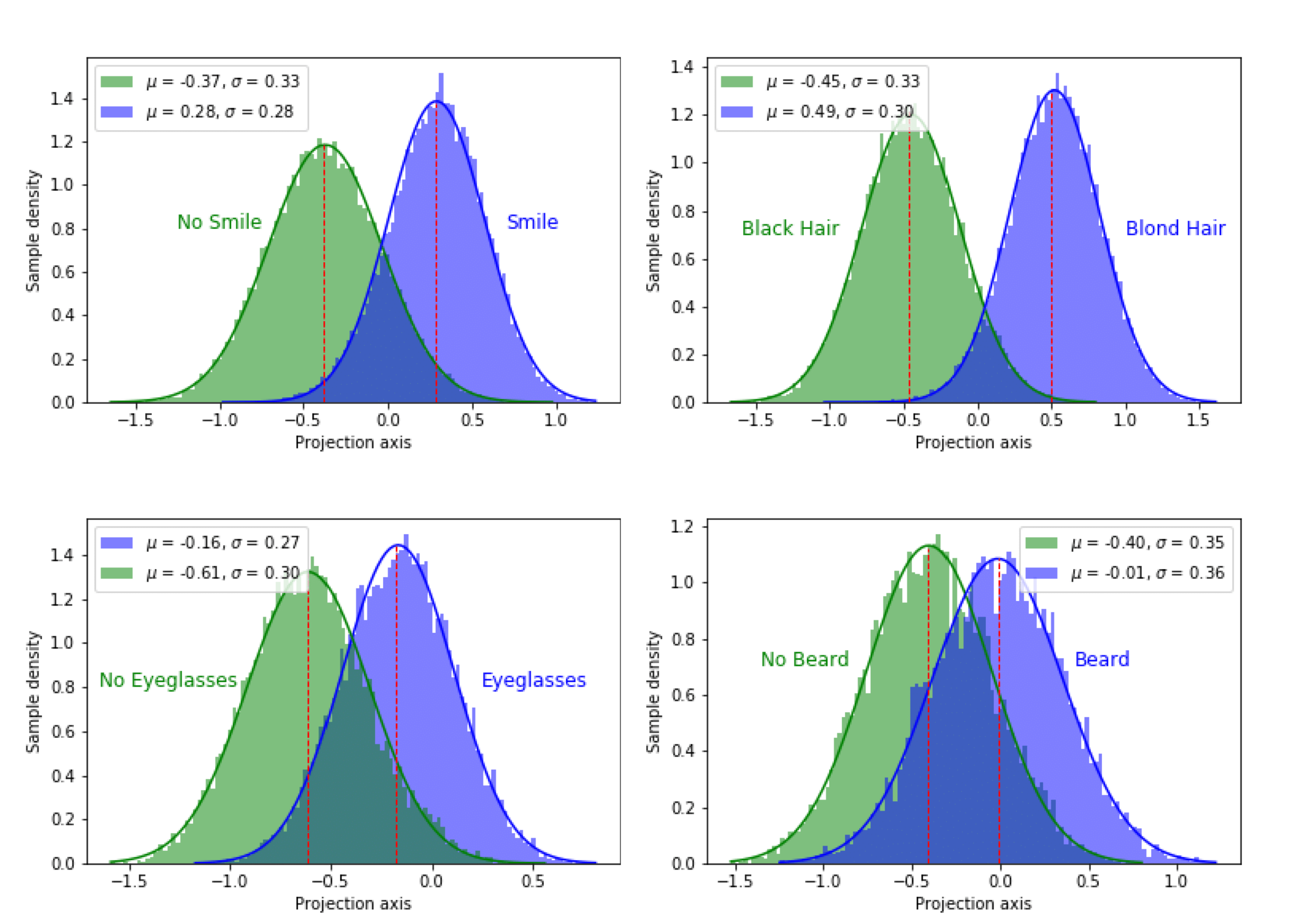}
	\caption{Projections on the latent attribute vector directions: top-left: smiling attribute, top-right: hair color, bottom-left: eyeglasses, bottom-right: beard attribute.}
	\label{fig:histogram}
\end{figure}

We computed attribute directions as we defined in Section \ref{sec:computing_latent_attribute}, using $50$ arbitrary reference images for each attribute. Then we computed an average attribute direction using these directions separately for each attribute. The projections on these directions for each attribute revealed two Normal densities, one for the neutral samples (displayed with green color) and one for the attributed samples (displayed with blue color), consistently for all the attributes (Fig. \ref{fig:histogram}). 

\begin{figure}[!h]
	\centering
	\includegraphics[width=\textwidth]{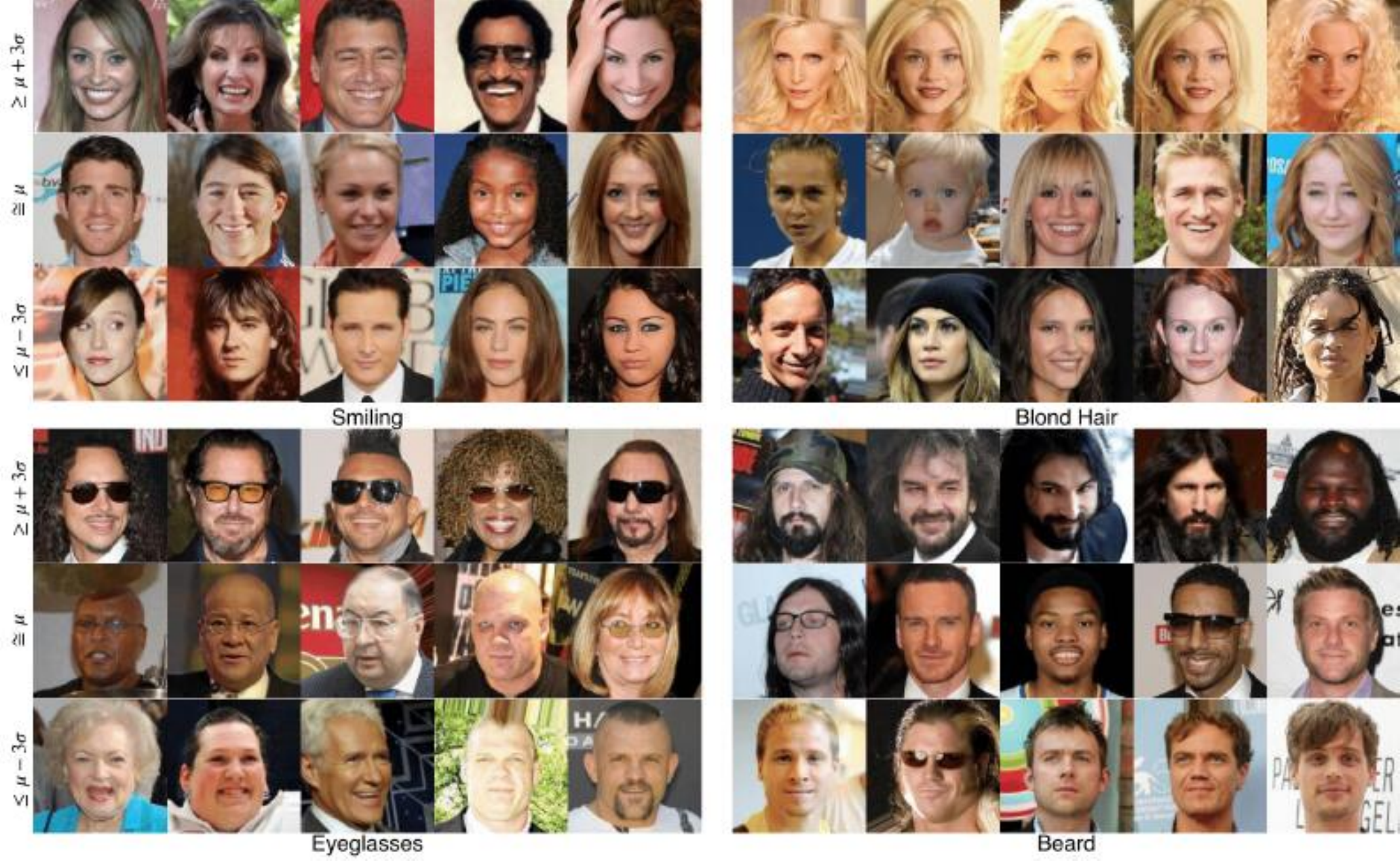}
	\caption{Sample images from the tails of Gaussian distributions for the four attributes.}
	\label{fig:gaussian_tails}
\end{figure}

We observed a set of randomly selected samples that are projected on different parts of the attribute distributions, i.e. the ones labeled as smile, blond hair, eyeglasses and beard distributions, in Fig. \ref{fig:gaussian_tails}. We intentionally selected samples that project on the left and right tails of the Normal distributions, i.e. beyond $\mu\pm3\sigma$, and around the mean. Note that, each image used in these blue densities are labeled with the related attribute in the CelebA dataset. We wondered the appearances of images on the left tail of these distributions. In Fig. \ref{fig:gaussian_tails}, for each attribute, the images on the top row are sampled from the right tail of the distributions, i.e. above $\mu+3\sigma$. We observe that the related attribute is explicitly visible in these images, which is as we expected since they are far away from the neutral distribution. The images on the middle row are sampled around the mean of the attribute distributions; they also have the related attributes, but not as strong as the ones on the right tail. The images on the third row are sampled from the left tail of the distribution, i.e. below $\mu-3\sigma$; these samples are close to the neutral image means, and as can be seen from the images the attributes are barely visible in these samples. Although these images are labeled with the corresponding attributes in the CelebA dataset, most of them do not visually contain the related attributes. For instance, the images labeled as smiling, do not smile at all (see the third row of the smiling attribute samples). It is similar for the other attributes as well. This is due to the labeling errors in CelebA dataset, which may affect the conditional GAN based models. However, our model is not affected, since we do not use any labels during training. These random selections from left, middle and right parts of the distributions show that the generator gradually encodes the related attributes through the computed attribute directions in a visually consistent manner. In our proposal, we utilize this structure for attribute editing (see Eq. \ref{equ:z_new}).

\begin{figure}[!h]
	\centering
	\includegraphics[width=\textwidth]{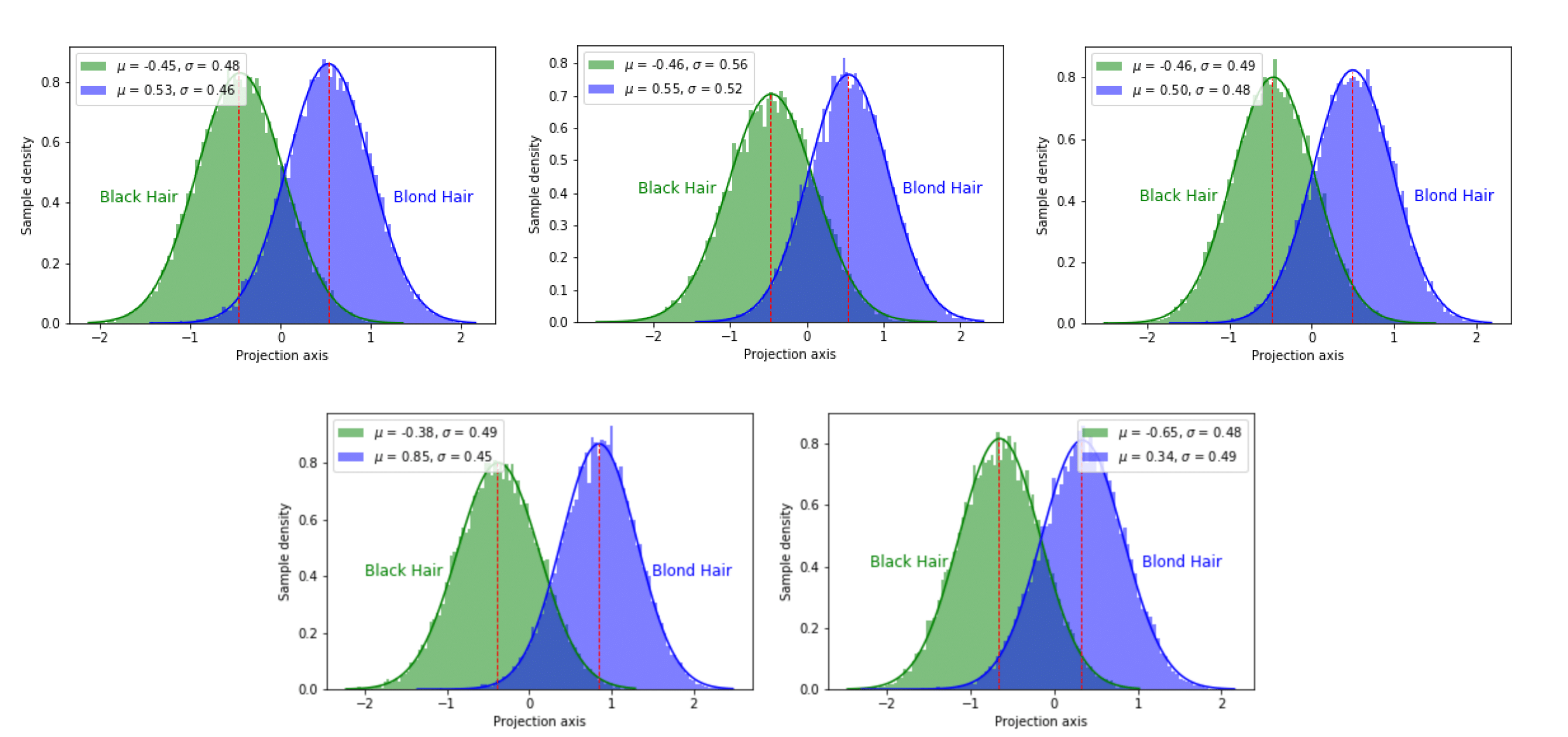}
	\caption{Sample distributions obtained using five random referance image directions with and without hair color attribute.}
	\label{fig:haircolor_attrib}
\end{figure}

Instead of the average direction, when we use any random reference image direction from the set of $50$ arbitrary reference images, we still observe two separate modalities after the projections, consistent with the distribution on the average attribute direction in Fig. \ref{fig:histogram}, with only slight changes in the means and standard deviations (Fig. \ref{fig:haircolor_attrib}). These observations show that a particular attribute can be coded by a range of linear directions that lie in close proximity; hence, the related attribute direction is not actually unique, there is a set of possible directions for each attribute. We provide visual examples to display the effect of arbitrary reference image selection in attribute editing further in Section \ref{sec:experiments_and_results}.

\section{Results and Discussion}
\label{sec:experiments_and_results}
This section provides comparisons of our CRG architecture with the related models designed for the concerned datasets and the results we obtained for image attribute editing at 128x128 pixel resolution using the CelebA dataset. 

\subsection{Image Reconstruction Experiments}
\label{sec:image_reconstruction_experiments}

\begin{figure}[!h]
	\centering
	\includegraphics[width=0.9\textwidth]{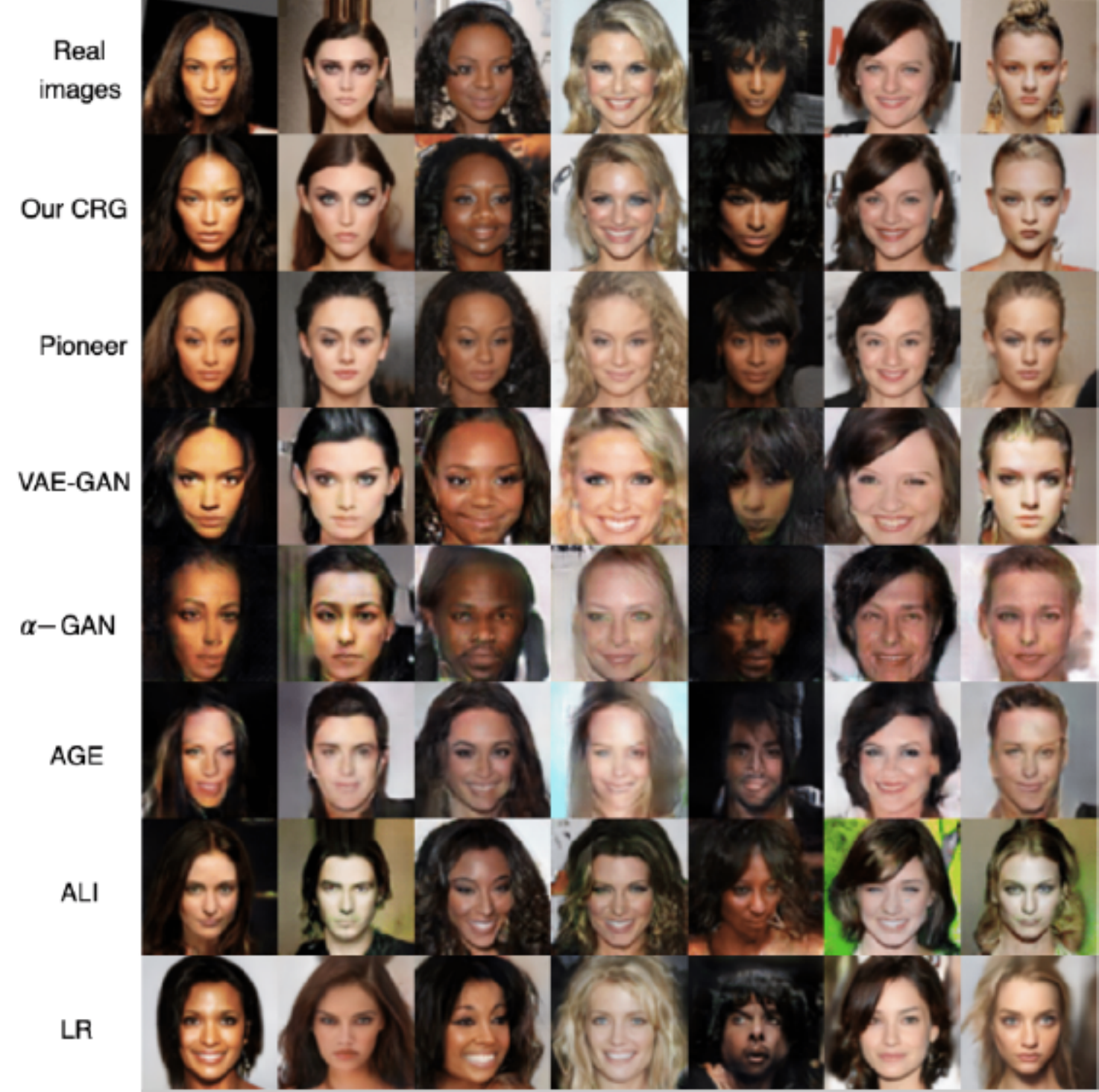}
	\caption{Sample reconstructions of the state of the art generative models. The images on the top row are used in the reconstructions. Pioneer, LR and our CRG model generate images in 128x128 pixel resolution, the other models generate images in 64x64 pixels.}
	\label{fig:Comparison_of_all_models}
\end{figure}

In this section, we compare the models that are designed for reconstruction of images using generative networks, i.e. Pioneer\footnote{\url{https://github.com/AaltoVision/pioneer}}, VAE-GAN\footnote{\url{https://github.com/PrateekMunjal/Autoencoding-beyond-pixels-using-a-learned-similarity-metric}}, $\alpha$-GAN\footnote{\url{https://github.com/PrateekMunjal/Variational-Approaches-for-Auto-Encoding-Generative-Adversarial-Networks}}\cite{rosca2017variational}, AGE\footnote{\url{https://github.com/DmitryUlyanov/AGE}}, ALI\footnote{\url{https://github.com/IshmaelBelghazi/ALI}}, LR models. We generated the models using their official pre-trained models; when we can not find official models, we either retrained the models from scratch or used a pre-trained model that is configured as in the original articles. Our CRG, Pioneer and LR models were trained using 30k images while the others were trained around 200k images. The comparison of the reconstructed images for real inputs are depicted in Fig. \ref{fig:Comparison_of_all_models}. As can be seen from the images, our CRG model reconstructs high-quality images that look more similar to the given inputs.

\begin{figure}[!h]
	\centering
	\includegraphics[width=0.9\textwidth]{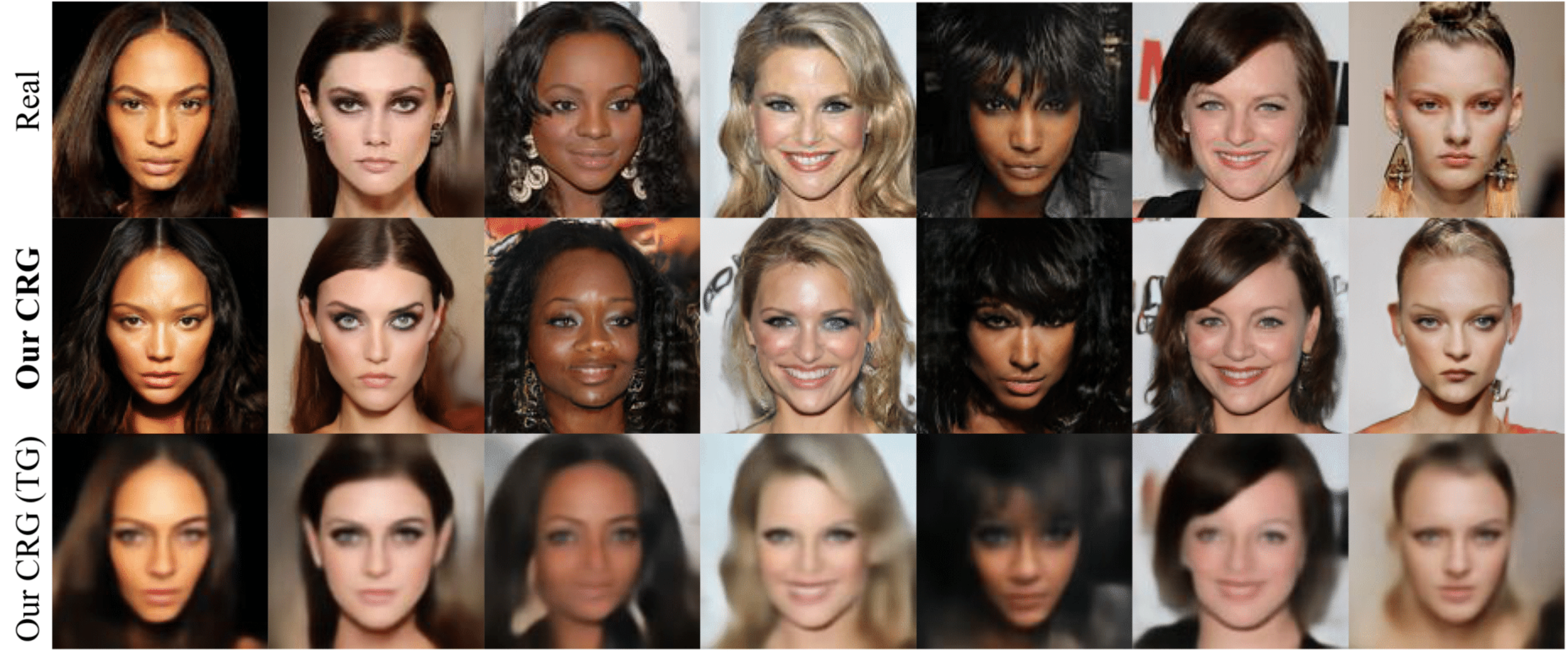}
	\caption{Comparison of real image reconstructions: (first row) real images that are used in the reconstruction, (second row) our CRG model (without generator training) and (third row) CRG model with generator training and encoder training simultaneously.}
	\label{fig:comparison_our_nonfixedGen}
\end{figure}

Moreover, we also retrained our model to see the effect of generator parameter updates while encoder training, i.e. we update the parameters of the generator as well during encoder training. We will refer to this model as CRG(TG) from this on. Fig. \ref{fig:comparison_our_nonfixedGen}, we provide sample reconstructions for our CRG and CRG(TG) methods. Since there is no discriminator in the CRG architecture, the generated images gradually lose their sharp features during minimization of the MAE and MSE loss functions in the generator parameter updates. Hence, we believe that it is better to first train a GAN in a domain which generates realistic images; then, without changing the generator parameters, train an encoder that learns the inverse of the generator using CRG.

We also computed the difference hash (dhash) \cite{niu2008overview}, perceptual hash (phash) \cite{niu2008overview} and wavelet hash (whash) \cite{whash} metrics to measure the similarity between the real and the reconstructed images. In these methods, images are initially converted into a fixed hash code. Then, the similarity between the original image hash code and the regenerated image hash code determines how the two images are similar. Table \ref{tab:CELEBA_dhash_phash_whash} provides the scores of these perceptual evaluation metrics, MAE and MSE for the selected models. The dhash, phash and whash metrics generate more perceptually relevant scores; when the reconstructed image is sharp and similar to the input image, they generate high scores. On the other hand, MAE and MSE provide averages of pixelwise differences. As can be seen from the Table, the perceptual scores of the CRG is better than all the other methods; yet MAE and MSE scores of the CRG are lower than the Pioneer. The blurred reconstructions of the Pioneer model results in less error, pixelwise. When we compute the MAE and MSE scores using our CRG(TG) reconstructions, we obtain $0.215$, $0.034$, respectively. Although this scores seem to be better than all the methods, the reconstructions of CRG(TG) are obviously very blurred compared to CRG. Hence, MAE and MSE metrics are not very useful for comparing the quality of different reconstructions. The perceptual metrics validate our visual judgments in Fig. \ref{fig:Comparison_of_all_models} that the CRG model reconstructions are better than the other model reconstructions for real images.	

%
%

\begin{table}[!h]
	\centering
	\caption{The similarities of real images and reconstructed images.}
	\label{tab:CELEBA_dhash_phash_whash}
	\begin{tabular}{llllll}
		\hline
		Model        & dhash          & phash          & whash	& MAE            & MSE         \\ \hline
		\textbf{CRG (TG)} & 0.846 & 0.792 & 0.855  &  \textbf{0.215}          &  \textbf{0.034} \\ \hline
		\textbf{CRG} & \textbf{0.871} & \textbf{0.848} & \textbf{0.892}  & 0.306          & 0.067 \\ \hline
		Pioneer \cite{heljakka2018pioneer}     & 0.852          & 0.804          & 0.864   & 0.261          & 0.051        \\ \hline
		VAE-GAN \cite{larsen2015autoencoding}      & 0.842          & 0.779          & 0.861    & 0.355 & 		0.095       \\ \hline
		$\alpha-$GAN \cite{rosca2017variational}    & 0.837          & 0.756          & 0.855   & 0.443          & 0.122       \\ \hline
		AGE \cite{ulyanov2018takes}         & 0.832          & 0.754          & 0.863        & 0.449          & 0.136    \\ \hline
		ALI \cite{dumoulin2016adversarially}           & 0.759          & 0.668          & 0.745   & 0.487          & 0.159        \\ \hline
		LR  \cite{donahue2016adversarial}         & 0.697          & 0.585          & 0.674       & 0.613          & 0.263    \\ \hline
	\end{tabular}
\end{table}

\begin{figure}[!h]
	\centering
	\includegraphics[width=.9\textwidth]{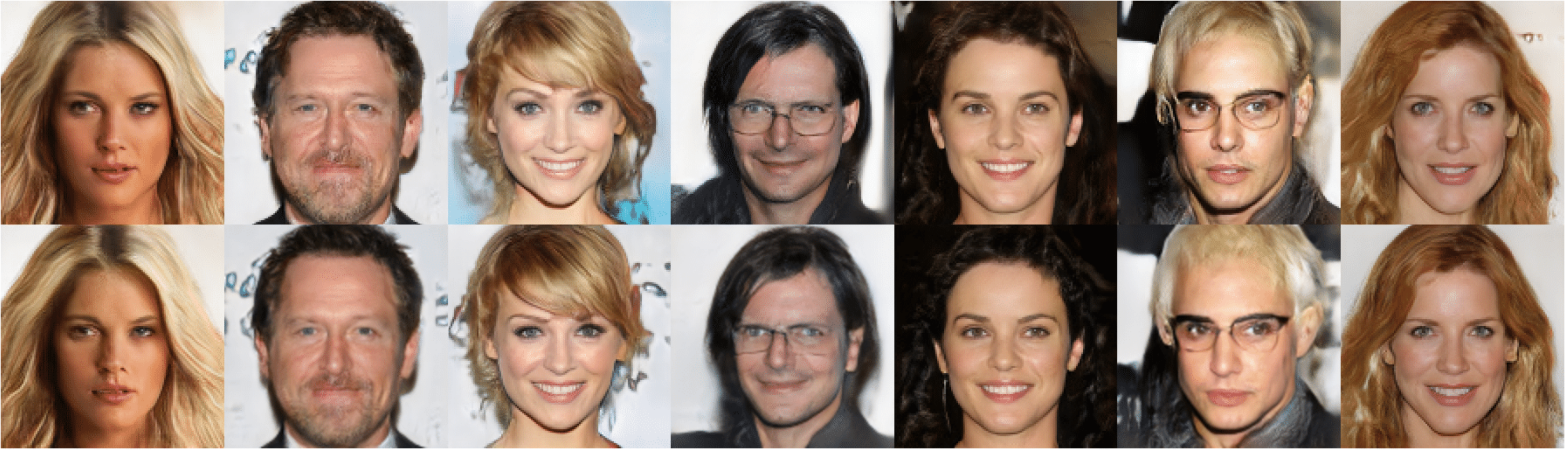}
	\caption{Top row: Randomly generated images, bottom row: reconstructed images using our model.}
	\label{fig:our_model_regenerate_images_fake}
\end{figure}

In order to change any attribute of a generated image, it is necessary to correctly reconstruct a generated image. Fig. \ref{fig:our_model_regenerate_images_fake} shows the reconstruction of some generated images using our model. As can be seen from the samples, the encoder successfully learns the inverse mapping of the generator.

\subsection{Image Attribute Editing Experiments}
\begin{figure}[!h]
	\centering
	\includegraphics[width=.7\textwidth]{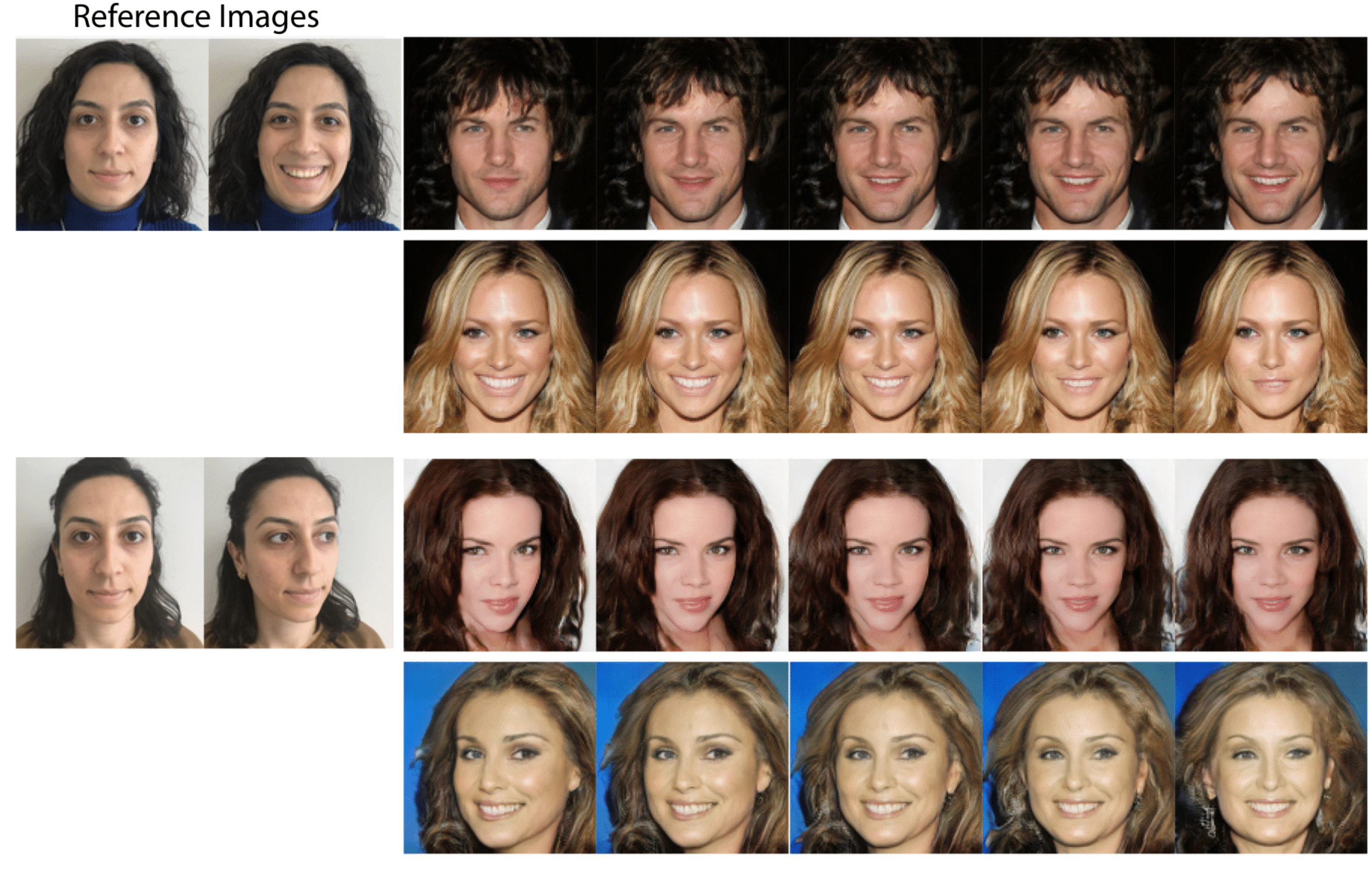}
	\caption{Smile and pose adjustments using arbitrary reference images. On the left side, reference image pairs are displayed. Multiple samplings of some generated images in the reference attribute direction are depicted on the right side.}
	\label{fig:smile_and_pos_attribute_separate_generated}
\end{figure}

In this section, we modify two attributes, the smile and the pose adjustment attributes, to demonstrate the attribute encoding of a source image using two arbitrary reference images. We show the results of manipulating the smile attribute in two directions and the pose adjustment of the source images with neutral and smiling gestures in Fig. \ref{fig:smile_and_pos_attribute_separate_generated}. As it is seen in the figures, with the exception of the interested attribute, other attributes of the source images are preserved substantially. These results are very promising for generating various synthetic images with desired attributes, which can be utilized in data augmentation for various problem domains. Using the proposed CRG training procedure, desired attributes can be injected to a source image without using a labeled dataset or conditional GANs architecture.

\begin{figure}[!h]
	\centering
	\includegraphics[width=.7\textwidth]{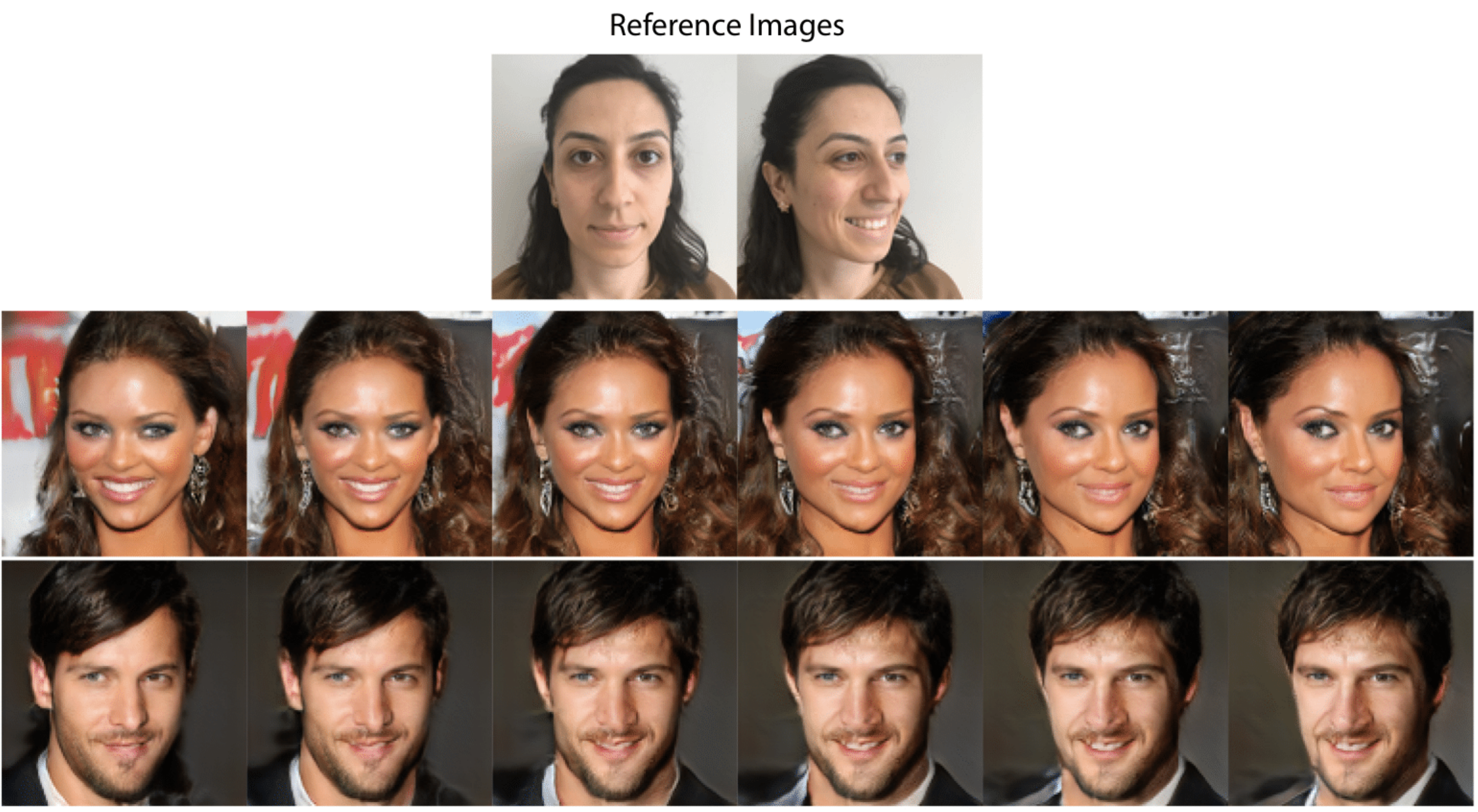}
	\caption{Simultaneously changing smile and pose adjustments using an arbitrary reference image pair. Top row: reference image pairs, Second and third row: multiple samplings in the reference attribute direction.}
	\label{fig:smile_and_pos_attribute_together_generated}
\end{figure}

We also made experiments on changing more than one attribute simultaneously, on purpose, while preserving the source images. In Fig. \ref{fig:smile_and_pos_attribute_together_generated}, we show manipulation of the pose adjustment and the smile attributes simultaneously. The results exemplify multiple attribute editing using again only one pair of reference images.

\begin{figure}[!h]
	\centering
	\includegraphics[width=.7\textwidth]{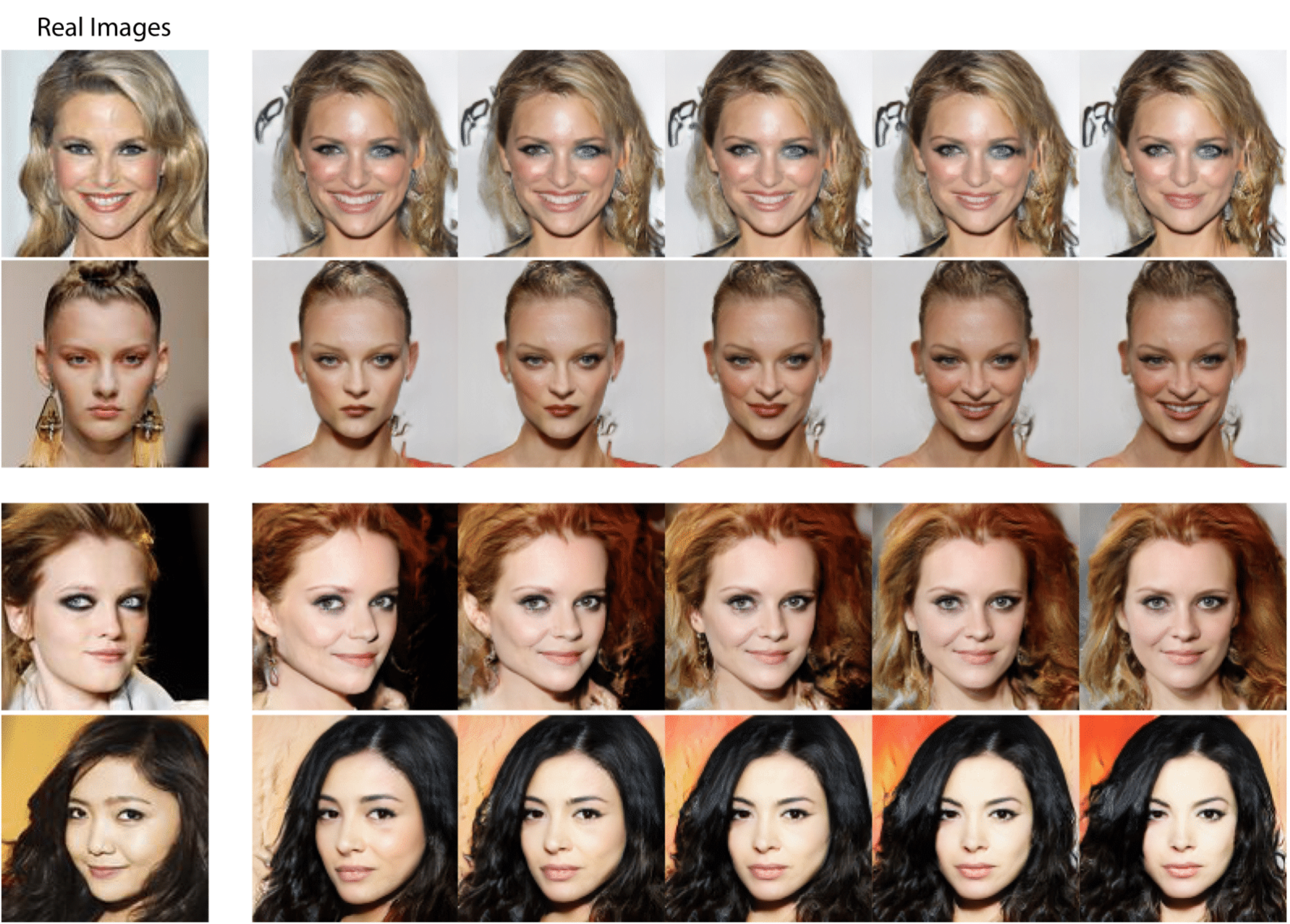}
	\caption{Smile attribute and pose adjustment on real images. The same smile and pose adjustment reference image pairs, depicted in Fig. \ref{fig:smile_and_pos_attribute_separate_generated}, are also used in these samples.}
	\label{fig:smile_and_pose_real_images}
\end{figure}

Although our aim is to modify only a subset of attributes of the generated images, we also tested the model's performance on real images. As shown in Fig. \ref{fig:smile_and_pose_real_images}, we found that the results in the real images are also promising. We believe that the difference between the original source and the reconstruction is mainly due to using insufficient number of samples during our GAN training, i.e. $30000$ real images. Generator may not know producing some novel attributes that it did not see before. For example, the earrings of the second sample in Fig. \ref{fig:smile_and_pose_real_images} is not produced since the generator has not seen enough samples with large earrings during training.  However, the hair styles, gesture characteristics and the pose of the input images are successfully captured in the reconstructed images. The accuracy of real image reconstruction can be increased by using more samples with many attribute variations during GAN training.

In the Supplementary Materials Section (Appendix \ref{supp_mats}), we provided additional image attribute editing results, in total 15 attributes, that have strong visual impact. These attributes are bald, bangs, hair length, brown hair, blond hair, black hair, gender, age, straight hair, wavy hair, sunglasses, smiling, eyeglasses, beard and skin color.

\subsection{Effect of arbitrary reference images in attribute coding}
\label{sec:Consistency_of_attribute_vectors}

\begin{figure}[!ht]
	\centering
	\includegraphics[width=\textwidth]{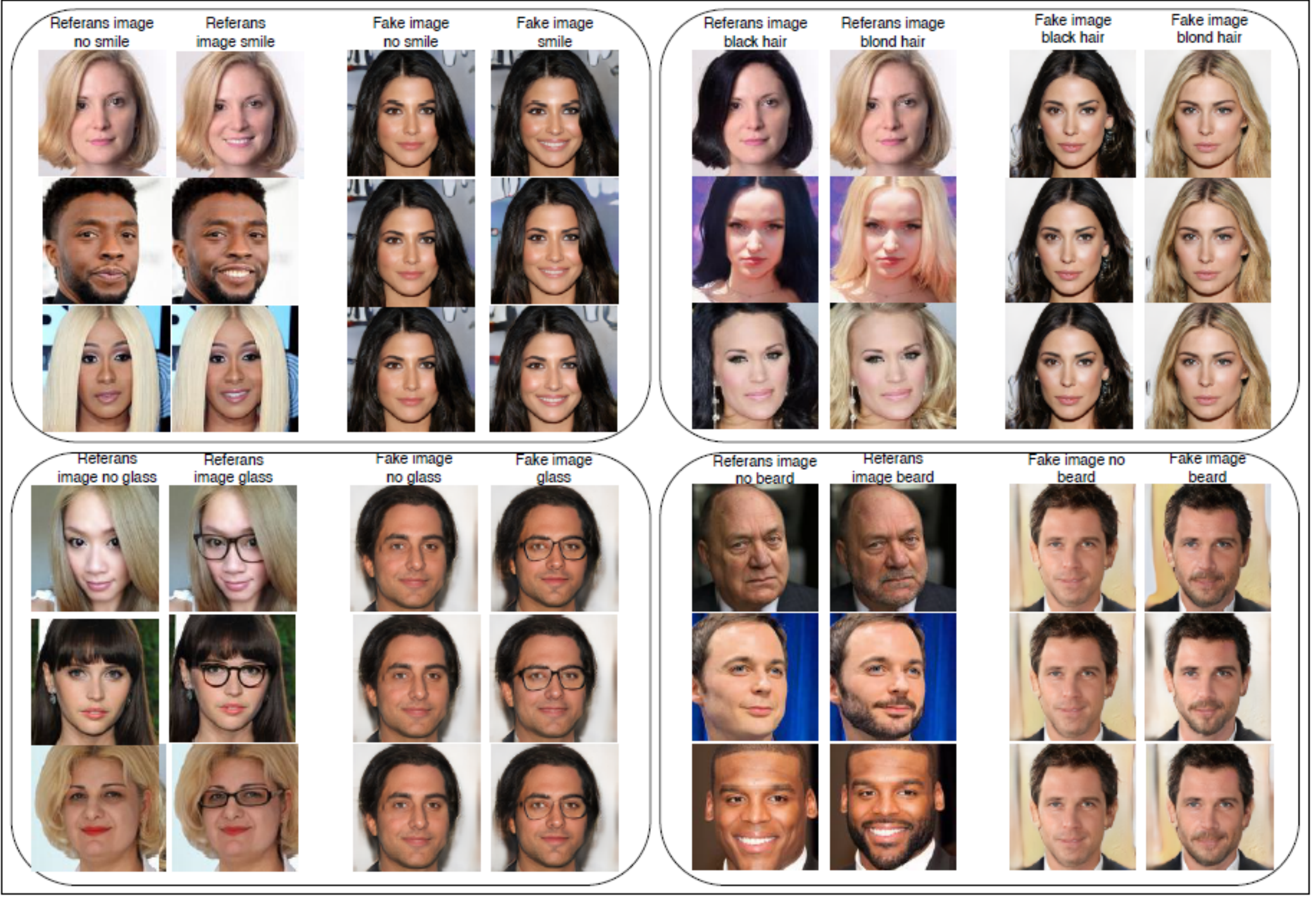}
	\caption{Attribute editing using different reference image pairs for the four attributes. Top left frame: smile, top right frame: hair color, bottom left frame: eyeglasses, bottom right frame: beard.}
	\label{fig:multiple_references}
\end{figure}

In this section, we elaborate the consistency of the attribute editings for an attribute using different reference image pairs. We use the same four attributes in our attribute direction analysis that we used in Section \ref{sec:Analysis_of_attrib_dir}, i.e. smile, hair color, eyeglasses and beard. The left side of each attribute frame in Fig. \ref{fig:multiple_references} shows three arbitrary reference image pairs that we prepared using real images. On the right side of each attribute window, we show the encoding of the attributes to the generated images with our method, using the attribute vector directions corresponding the reference image pairs on the left. While using Eq. \ref{equ:z_new}, we set $k$ to $2$ for all the attribute editings in these samples. Although the reference images in attribute direction computations are quite different, since the reference images are consistent in the neutral and attributed images in all but the selected attributes, the computed directions are consistent and similar. When the generated images are carefully examined, we realize some minor differences in the generated images due to the differences in the reference images, such as hair boundaries, backgrounds etc. Such minor differences are expected as a result of small variations in the computed directions. These results show that arbitrary reference image pairs can be used to encode an attribute to an input image without changing the primary visual features on the input image. 

\subsection{Effective range for $k$}
\label{sec:Range_for_k}

\begin{figure}[!h]
	\centering
	\includegraphics[width=\textwidth]{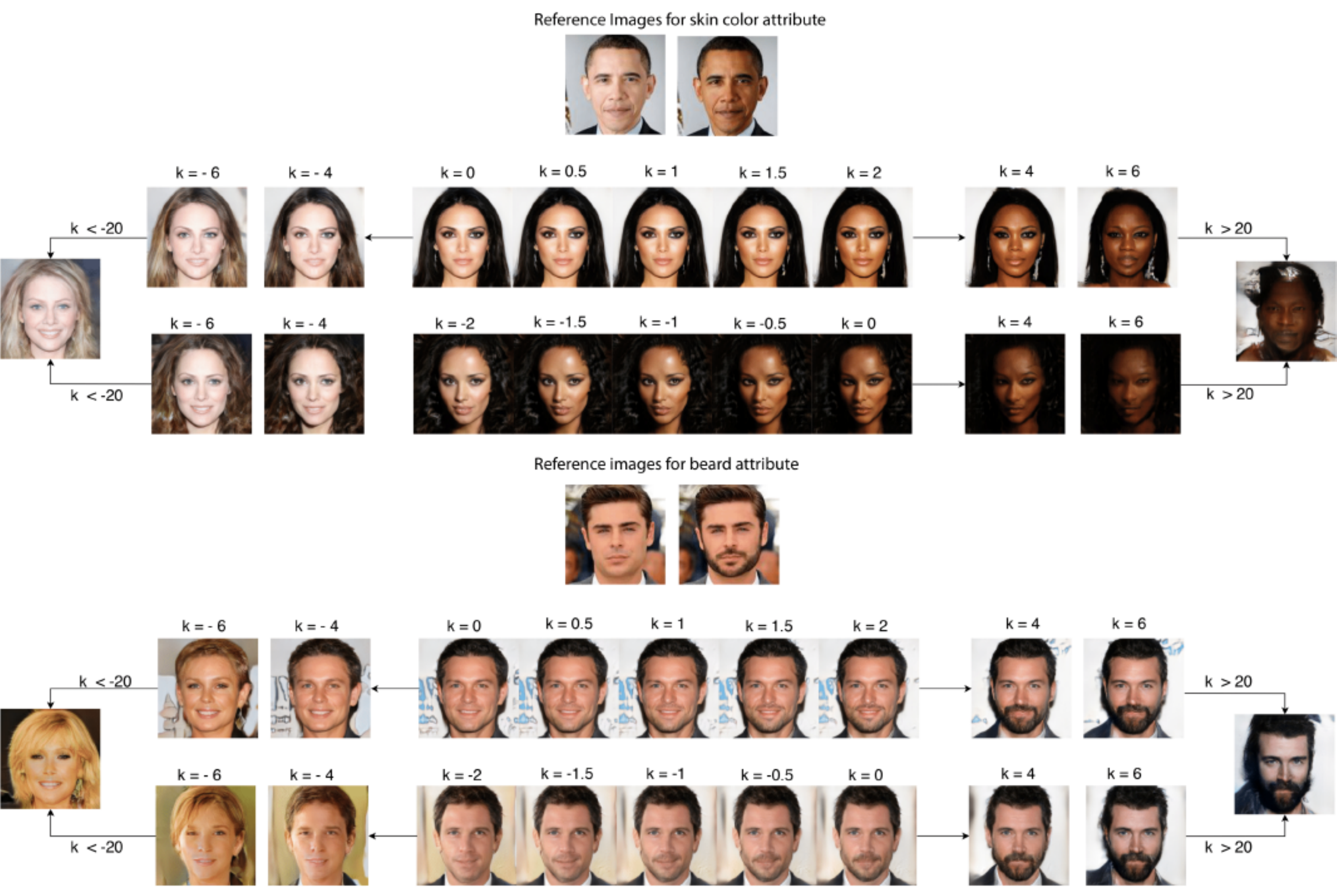}
	\caption{Examination of the effect of the k parameter for a range of values. Top: change of skin color attribute for a range of k, bottom: change of beard attribute. The reference image pair are given above each sample. For both attributes, k = 0 represents the start image.}
	\label{fig:boundary_of_k}
\end{figure}

In order to understand how the images in an attribute direction change, we performed a set of observations by changing $k$, in Eq. \ref{equ:z_new}, in positive and negative directions for a large range of values. The results for two of the selected attributes, namely the skin color attribute and beard, are depicted in Fig. \ref{fig:boundary_of_k}. As can easily be seen in skin color attribute samples, the related attribute gradually increase/decrease in positive/negative directions, as we expected. In the local neighborhoods of a generated image, the attributes other than the edited, i.e. skin color and beard in these examples, are preserved. When we continue incrementing/decrementing $k$ in the attribute vector directions, images gradually change to different faces. Hence, the editing without changing the primary face features needs to be done in the local neighborhood of a given image. When the change on the axis is implemented with high margins from the original image, by means of using large magnitudes of $k$ values, e.g. $k>20$ or $k<-20$, independent of the starting image, all images converge to a particular pair of face image on the left and right ends. Indeed, there is no end on this axis, yet after some point the interpolation on the axis does not produce new images. This is not surprising, since at the very far ends of these axes in the latent space, there are almost no observed samples by the generator due to the z sampling we use during GAN training; each component of z is sampled from a Normal distribution with $\mu_{z_i}=0$ and $\sigma_{z_i}=1$. When we go very far, i.e. a lot more than $\mu_{z_i}\pm3\sigma_{z_i}$ in the components of $z$, i.e. $z_i$, that are effected by this attribute direction, the probability of change in the generated image get very close to zero. We observe the same issue with both attributes, skin color and beard, in a similar fashion (Fig. \ref{fig:boundary_of_k}). 

In general, we observed that when we go through positive or negative direction in an attribute axis with small steps, the other input features are kept very similar to the original input when $k<2$. The sequence of images in the middle sections of the Fig. \ref{fig:boundary_of_k} are sampled with  $k=0.5$ to better display this observation.

In order to make a more conservative analysis to determine the ranges of $k$, we can utilize the sample distributions that we presented in Section \ref{sec:Analysis_of_attrib_dir}. Utilizing such a distribution for an attribute may be helpful to define a more systematic way for determining the boundaries of $k$. What is meant with conservative analysis is that, we want to find the effective boundaries of $k$ for an attribute so that the essential face attributes of a given image is preserved while editing only the selected attribute. As we presented in Section \ref{sec:Analysis_of_attrib_dir}, we obtain two Normal densities in each attribute direction when we project real images in the computed directions; one for the neutral images, i.e. neutral with respect to the selected attribute, and one for the attributed images. The mean and standard deviations of each distribution, for different attributes, can guide us to analyze and define the safe rages for $k$. As we have already shown in Fig. \ref{fig:gaussian_tails}, the samples beyond $3\sigma$ on the left tails of the attribute distributions contain samples with barely visible attributes, while the right tails contain images with very exaggerated attributes. Hence, we believe that going beyond $\mu\pm3\sigma$ on an attribute axis may create some noticeable changes in the given input facial attributes, which we want to preserve during editing. In this view, for any z in the latent space, we can formulate the range of $k$ by computing its projected distances to $\mu\pm3\sigma$ on the attribute axis. We then use the values in this range for safely editing only the related attributes of an image.

\begin{figure}[!h]
	\centering
	\includegraphics[width=0.7\textwidth]{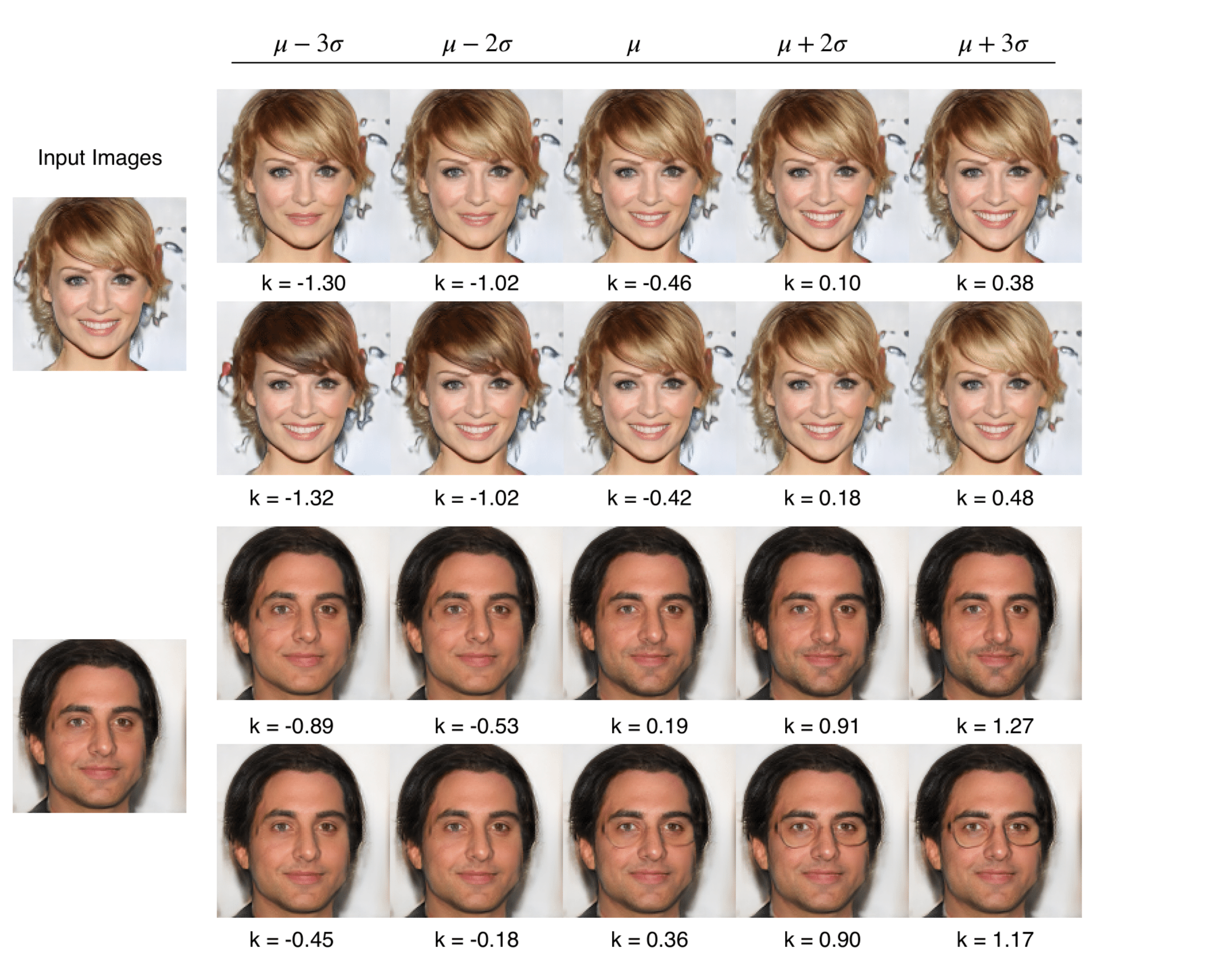}
	\caption{Left column: sample input images, right column: edited images using different values for $k$ for four attributes.}
	\label{fig:sigma_analysis_for_k}
\end{figure}

In order to show the idea with an example, we selected two sample images and edit them using the set of attributes that we analyzed in Section \ref{sec:Analysis_of_attrib_dir} (in Fig. \ref{fig:sigma_analysis_for_k}). We first compute the z values of these images using our Encoder. Then, we compute the projections of their z vectors on the related attribute directions, separately for each attribute, i.e. $a$. Then we compute the projected distances of the images to the $\mu_a\pm3\sigma_a$ of the attribute distributions, i.e. subscript $a$ represents the relevant attribute. We set these distances as the bounds for $k$ values in our computations in these examples. In order to observe the generated images for the computed range, we set $k$ in such a way that it coincides with $\mu_a$,  $\mu_a\pm2\sigma_a$ and $\mu_a\pm3\sigma_a$ points on the attribute axes. Using the original z values of both images and our encoder, we edit the images using our simple formula, Eq. \ref{equ:z_new}, with the determined $k$ values for the related four attributes. The resultant images are depicted in Fig. \ref{fig:sigma_analysis_for_k}. As can be seen from the images, when $k$ is set to the values between $\mu\pm3\sigma$ of the attribute distributions, the attribute editing  successfully preserves other facial attributes while editing only the desired attributes. When we look at the distribution parameters, the average distance of a neutral image to the right tail (i.e. $\mu+3\sigma$) of the attributed distribution changes between $1.26$ to $1.84$ for the selected four attributes. However, this range becomes smaller if the input image already has the attribute, such as the first image, which has blond hair and smile attributes already. In such cases, the projection of the original input is already very close to the attribute mean on the attribute axis, which results in very small changes in the target attribute. These values are consistent with our observations; in our experiments, for editing some attributes in neutral images, we observed that $k=2$ usually works without changing the other facial attributes. Since the range of the selected attribute distributions are less than $2$ in these experiments, the resultant images in Fig. \ref{fig:sigma_analysis_for_k} preserves other facial attributes of the image while editing.

\section{Conclusion and Future Works}
\label{sec:conclusion_and_future_works}
We introduced a novel CRG architecture that is effective learning the inverse of a given generator using cyclic error minimization without supervision. The proposed architecture can be used to compute the accurate latent representations of generated images for attribute editing. The attribute editing can be performed by providing a pair of real images that contains a source image with and without desired attributes. The rate of manipulation of the input images can be controlled dynamically using only one parameter. The set of attributes for editing is also dynamic and determined via reference images. The results of the experiments show that the quality of the image reconstructions is more successful than the state-of-the-art models. Although the reconstruction of the generated images is very successful, the reconstruction performance of the model with the real images may be improved by using more training data with increased attribute variation. As a future work, in addition to increasing training data variation, we are also planning to train the encoder progressively, in parallel with generator training, to increase the reconstruction performance of the real images. Moreover, we are planning to make more comprehensive analysis on the latent spaces of various GAN generators using our CRG method.

\section*{Acknowledgements}
	This research is funded by Ankara University (Scientific Research Projects Grant, grant id: 18L0443010). The numerical calculations reported in this paper were partially performed at TUBITAK ULAKBIM, High Performance and Grid Computing Center (TRUBA resources). We would like to thank the anonymous reviewers for their valuable suggestions and comments. We also thank Dr. Kai Dierkes for the helpful discussions and comments during the revision process.

\bibliographystyle{ieeetr}
\footnotesize \bibliography{CRGRefs}

\begin{thebibliography}{10}

\bibitem{goodfellow2014generative}
I.~Goodfellow, J.~Pouget-Abadie, M.~Mirza, B.~Xu, D.~Warde-Farley, S.~Ozair,
  A.~Courville, and Y.~Bengio, ``Generative adversarial nets,'' in {\em
  Advances in neural information processing systems}, pp.~2672--2680, 2014.

\bibitem{kingma2018glow}
D.~P. Kingma and P.~Dhariwal, ``Glow: Generative flow with invertible 1x1
  convolutions,'' in {\em Advances in Neural Information Processing Systems},
  pp.~10215--10224, 2018.

\bibitem{kingma2016improved}
D.~P. Kingma, T.~Salimans, R.~Jozefowicz, X.~Chen, I.~Sutskever, and
  M.~Welling, ``Improved variational inference with inverse autoregressive
  flow,'' in {\em Advances in neural information processing systems},
  pp.~4743--4751, 2016.

\bibitem{oord2016pixel}
A.~v.~d. Oord, N.~Kalchbrenner, and K.~Kavukcuoglu, ``Pixel recurrent neural
  networks,'' {\em arXiv preprint arXiv:1601.06759}, 2016.

\bibitem{radford2015unsupervised}
A.~Radford, L.~Metz, and S.~Chintala, ``Unsupervised representation learning
  with deep convolutional generative adversarial networks,'' {\em arXiv
  preprint arXiv:1511.06434}, 2015.

\bibitem{rosca2017variational}
M.~Rosca, B.~Lakshminarayanan, D.~Warde-Farley, and S.~Mohamed, ``Variational
  approaches for auto-encoding generative adversarial networks,'' {\em arXiv
  preprint arXiv:1706.04987}, 2017.

\bibitem{liu2015deep}
Z.~Liu, P.~Luo, X.~Wang, and X.~Tang, ``Deep learning face attributes in the
  wild,'' in {\em Proceedings of the IEEE international conference on computer
  vision}, pp.~3730--3738, 2015.

\bibitem{mirza2014conditional}
M.~Mirza and S.~Osindero, ``Conditional generative adversarial nets,'' {\em
  arXiv preprint arXiv:1411.1784}, 2014.

\bibitem{antipov2017face}
G.~Antipov, M.~Baccouche, and J.-L. Dugelay, ``Face aging with conditional
  generative adversarial networks,'' in {\em 2017 IEEE International Conference
  on Image Processing (ICIP)}, pp.~2089--2093, IEEE, 2017.

\bibitem{perarnau2016invertible}
G.~Perarnau, J.~Van De~Weijer, B.~Raducanu, and J.~M. {\'A}lvarez, ``Invertible
  conditional gans for image editing,'' {\em arXiv preprint arXiv:1611.06355},
  2016.

\bibitem{jaiswal2018bidirectional}
A.~Jaiswal, W.~AbdAlmageed, Y.~Wu, and P.~Natarajan, ``Bidirectional
  conditional generative adversarial networks,'' in {\em Asian Conference on
  Computer Vision}, pp.~216--232, Springer, 2018.

\bibitem{choi2018stargan}
Y.~Choi, M.~Choi, M.~Kim, J.-W. Ha, S.~Kim, and J.~Choo, ``Stargan: Unified
  generative adversarial networks for multi-domain image-to-image
  translation,'' in {\em Proceedings of the IEEE Conference on Computer Vision
  and Pattern Recognition}, pp.~8789--8797, 2018.

\bibitem{he2019attgan}
Z.~He, W.~Zuo, M.~Kan, S.~Shan, and X.~Chen, ``Attgan: Facial attribute editing
  by only changing what you want,'' {\em IEEE Transactions on Image
  Processing}, 2019.

\bibitem{lipton2017precise}
Z.~C. Lipton and S.~Tripathi, ``Precise recovery of latent vectors from
  generative adversarial networks,'' {\em arXiv preprint arXiv:1702.04782},
  2017.

\bibitem{creswell2018inverting}
A.~Creswell and A.~A. Bharath, ``Inverting the generator of a generative
  adversarial network,'' {\em IEEE transactions on neural networks and learning
  systems}, 2018.

\bibitem{larsen2015autoencoding}
A.~B.~L. Larsen, S.~K. S{\o}nderby, H.~Larochelle, and O.~Winther,
  ``Autoencoding beyond pixels using a learned similarity metric,'' {\em arXiv
  preprint arXiv:1512.09300}, 2015.

\bibitem{donahue2016adversarial}
J.~Donahue, P.~Kr{\"a}henb{\"u}hl, and T.~Darrell, ``Adversarial feature
  learning,'' {\em arXiv preprint arXiv:1605.09782}, 2016.

\bibitem{dumoulin2016adversarially}
V.~Dumoulin, I.~Belghazi, B.~Poole, O.~Mastropietro, A.~Lamb, M.~Arjovsky, and
  A.~Courville, ``Adversarially learned inference,'' {\em arXiv preprint
  arXiv:1606.00704}, 2016.

\bibitem{luo2017learning}
J.~Luo, Y.~Xu, C.~Tang, and J.~Lv, ``Learning inverse mapping by autoencoder
  based generative adversarial nets,'' in {\em International Conference on
  Neural Information Processing}, pp.~207--216, Springer, 2017.

\bibitem{heljakka2018pioneer}
A.~Heljakka, A.~Solin, and J.~Kannala, ``Pioneer networks: Progressively
  growing generative autoencoder,'' {\em arXiv preprint arXiv:1807.03026},
  2018.

\bibitem{ulyanov2018takes}
D.~Ulyanov, A.~Vedaldi, and V.~Lempitsky, ``It takes (only) two: Adversarial
  generator-encoder networks,'' in {\em Thirty-Second AAAI Conference on
  Artificial Intelligence}, 2018.

\bibitem{karras2017progressive}
T.~Karras, T.~Aila, S.~Laine, and J.~Lehtinen, ``Progressive growing of gans
  for improved quality, stability, and variation,'' {\em arXiv preprint
  arXiv:1710.10196}, 2017.

\bibitem{arjovsky2017wasserstein}
M.~Arjovsky, S.~Chintala, and L.~Bottou, ``Wasserstein generative adversarial
  networks,'' in {\em International Conference on Machine Learning},
  pp.~214--223, 2017.

\bibitem{gulrajani2017improved}
I.~Gulrajani, F.~Ahmed, M.~Arjovsky, V.~Dumoulin, and A.~C. Courville,
  ``Improved training of wasserstein gans,'' in {\em Advances in Neural
  Information Processing Systems}, pp.~5767--5777, 2017.

\bibitem{kodali2017convergence}
N.~Kodali, J.~Abernethy, J.~Hays, and Z.~Kira, ``On convergence and stability
  of gans,'' {\em arXiv preprint arXiv:1705.07215}, 2017.

\bibitem{mao2017least}
X.~Mao, Q.~Li, H.~Xie, R.~Y. Lau, Z.~Wang, and S.~Paul~Smolley, ``Least squares
  generative adversarial networks,'' in {\em Proceedings of the IEEE
  International Conference on Computer Vision}, pp.~2794--2802, 2017.

\bibitem{berthelot2017began}
D.~Berthelot, T.~Schumm, and L.~Metz, ``Began: Boundary equilibrium generative
  adversarial networks,'' {\em arXiv preprint arXiv:1703.10717}, 2017.

\bibitem{miyato2018spectral}
T.~Miyato, T.~Kataoka, M.~Koyama, and Y.~Yoshida, ``Spectral normalization for
  generative adversarial networks,'' {\em arXiv preprint arXiv:1802.05957},
  2018.

\bibitem{salimans2016improved}
T.~Salimans, I.~Goodfellow, W.~Zaremba, V.~Cheung, A.~Radford, and X.~Chen,
  ``Improved techniques for training gans,'' in {\em Advances in neural
  information processing systems}, pp.~2234--2242, 2016.

\bibitem{lucic2018gans}
M.~Lucic, K.~Kurach, M.~Michalski, S.~Gelly, and O.~Bousquet, ``Are gans
  created equal? a large-scale study,'' in {\em Advances in neural information
  processing systems}, pp.~700--709, 2018.

\bibitem{karras2018style}
T.~Karras, S.~Laine, and T.~Aila, ``A style-based generator architecture for
  generative adversarial networks,'' {\em arXiv preprint arXiv:1812.04948},
  2018.

\bibitem{huang2017arbitrary}
X.~Huang and S.~Belongie, ``Arbitrary style transfer in real-time with adaptive
  instance normalization,'' in {\em Proceedings of the IEEE International
  Conference on Computer Vision}, pp.~1501--1510, 2017.

\bibitem{doganyahya}
Y.~Dogan and H.~Yalim~Keles, ``Stability and diversity in generative
  adversarial networks,'' in {\em 2019 27nd Signal Processing and
  Communications Applications Conference (SIU)}, IEEE, 2019.

\bibitem{zhang2018self}
H.~Zhang, I.~Goodfellow, D.~Metaxas, and A.~Odena, ``Self-attention generative
  adversarial networks,'' {\em arXiv preprint arXiv:1805.08318}, 2018.

\bibitem{heusel2017gans}
M.~Heusel, H.~Ramsauer, T.~Unterthiner, B.~Nessler, and S.~Hochreiter, ``Gans
  trained by a two time-scale update rule converge to a local nash
  equilibrium,'' in {\em Advances in Neural Information Processing Systems},
  pp.~6626--6637, 2017.

\bibitem{zeiler2014visualizing}
M.~D. Zeiler and R.~Fergus, ``Visualizing and understanding convolutional
  networks,'' in {\em European conference on computer vision}, pp.~818--833,
  Springer, 2014.

\bibitem{simonyan2014very}
K.~Simonyan and A.~Zisserman, ``Very deep convolutional networks for
  large-scale image recognition,'' {\em arXiv preprint arXiv:1409.1556}, 2014.

\bibitem{tieleman2012lecture}
T.~Tieleman and G.~Hinton, ``Lecture 6.5-rmsprop: Divide the gradient by a
  running average of its recent magnitude,'' {\em COURSERA: Neural networks for
  machine learning}, vol.~4, no.~2, pp.~26--31, 2012.

\bibitem{niu2008overview}
X.-m. Niu and Y.-h. Jiao, ``An overview of perceptual hashing,'' {\em Acta
  Electronica Sinica}, vol.~36, no.~7, pp.~1405--1411, 2008.

\bibitem{whash}
D.~Petrov, ``{Wavelet image hash in Python}.''
  \url{https://fullstackml.com/wavelet-image-hash-in-python-3504fdd282b5/},
  2016.
\newblock [Online; accessed 19-June-2019].

\end{thebibliography}

\clearpage
 
\begin{appendices}
\renewcommand\thefigure{\thesection.\arabic{figure}}    
\section{Supplementary Materials}
\label{supp_mats}
In our method, the two reference images are designed using usually the same person, with and without an attribute. Hence, it is difficult to define reference image pairs for gender encoding; since many of the other attributes may change in the resultant image after editing when we just provide a random female and male faces as reference image pairs. Therefore, in order to create the reference images, we artificially added a beard to a female face, instead of providing a random male face (in female-to-male/male-to-female editing).

\setcounter{figure}{0}

\begin{figure}[!h]
	\centering
	\includegraphics[width=0.79\textwidth]{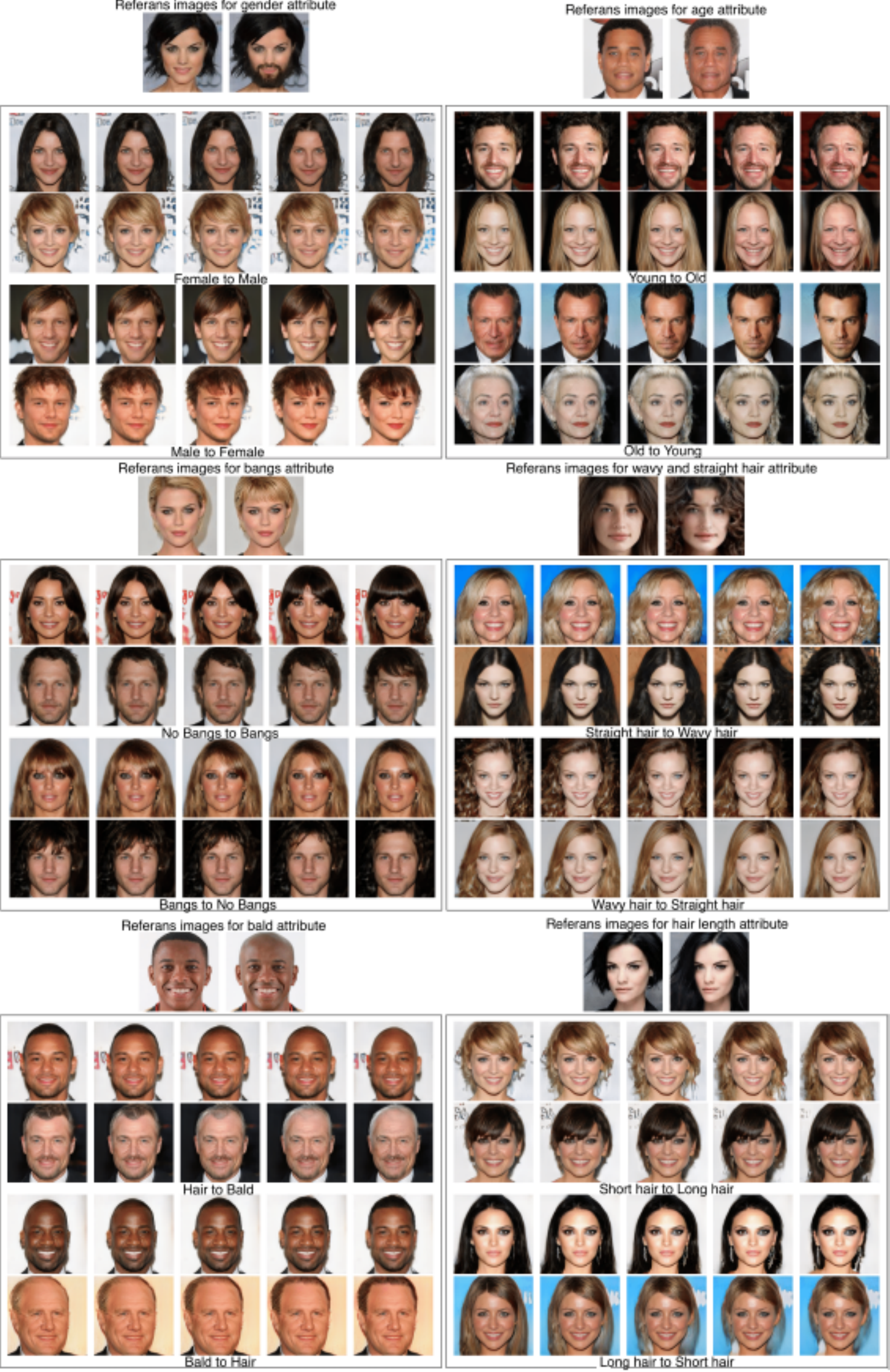}
	\caption{Additional attribute editing results for gender, age, bangs, wavy hair, straight hair, bald and hair length attributes. Attribute editing was done from left to right for all attributes. Best viewed in electronic format.}
	\label{fig:additional_results_1}
\end{figure}

\begin{figure}[!h]
	\centering
	\includegraphics[width=0.9\textwidth]{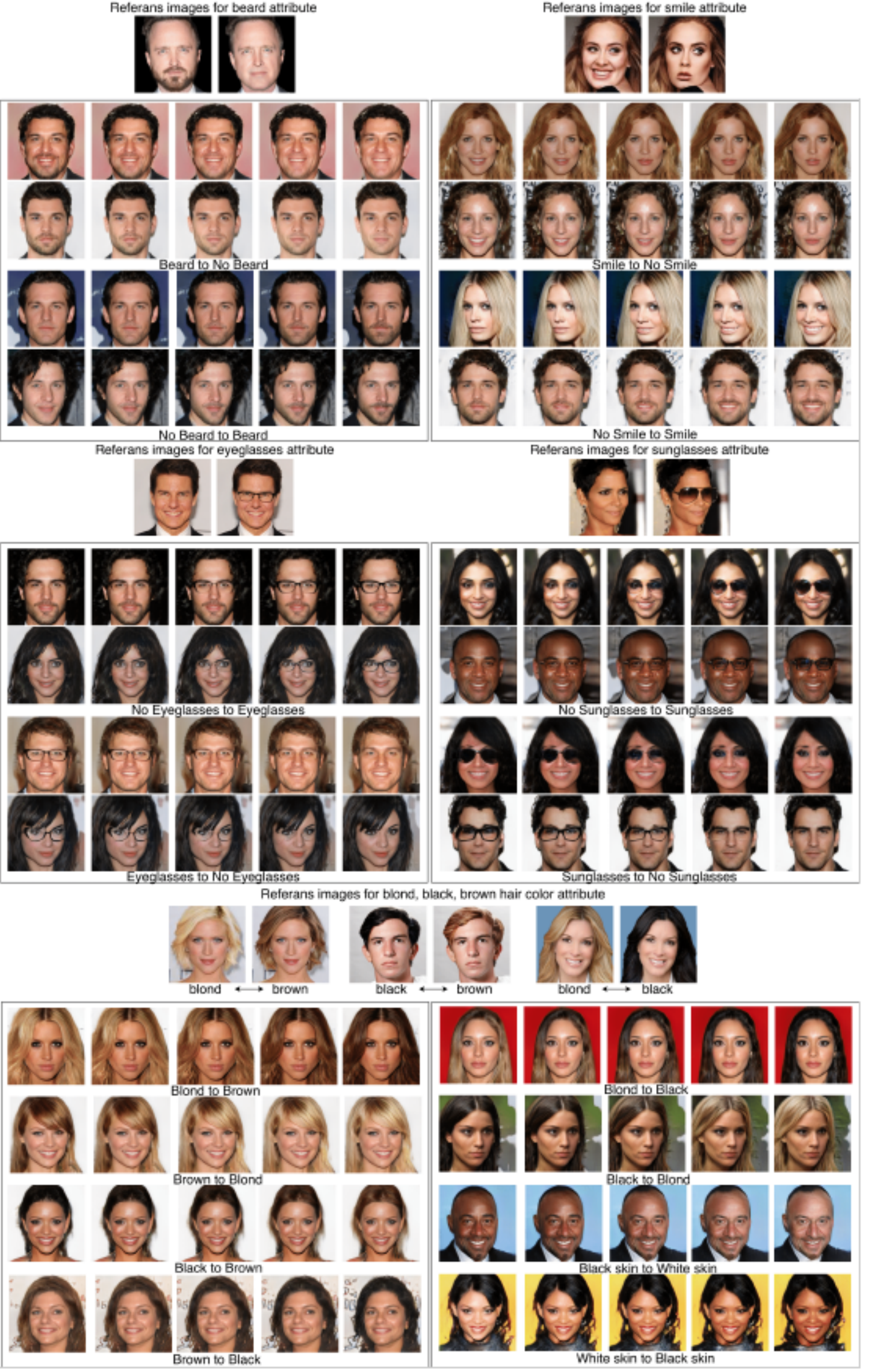}
	\caption{Additional attribute editing results for beard, smile, eyeglasses, sunglasses, blond hair, black hair, brown hair and skin color attributes. Attribute editing was done from left to right for all attributes. For skin color attribute, reference image pair in Fig \ref{fig:boundary_of_k} was used. Best viewed in electronic format.}
	\label{fig:additional_results_2}
\end{figure}

\end{appendices}

\end{document}